\documentclass{article}

\usepackage{arxiv}

\usepackage[utf8]{inputenc} 
\usepackage[T1]{fontenc}    
\usepackage{hyperref}       
\usepackage{url}            
\usepackage{booktabs}       
\usepackage{amsfonts}       
\usepackage{nicefrac}       
\usepackage{microtype}      
\usepackage{lipsum}
\usepackage{graphicx}

\title{Comparison of synthetic dataset generation methods for medical intervention rooms using medical clothing detection as an example}

\author{
	Patrick Schülein$\dagger$ 
	\and
	Hannah Teufel$\ddagger$
	\and 
	Ronja Vorpahl$\dagger$
	\and
	Indira Emter$\dagger$
	\and
	Yannick Bukschat$\dagger$
	\and
	Marcus Pfister$\diamond$
	\and
	Anke Siebert$\diamond$
	\and
	Nils Rathmann$\wedge$
	\and
	Steffen Diehl$\wedge$
	\and
	Marcus Vetter$\dagger$
	\\ 
	\\
	$\dagger$ ESM-Institute\\
	University of Applied Sciences Mannheim, Mannheim\\\\
	$\wedge$ Clinic for Radiology and Nuclear Medicine\\ Heidelberg University\\\\
	$\diamond$ Siemens Healthcare GmbH Forchheim, Germany \\\\
	$\ddagger$ During grad-student work at the ESM-Institute.\\Not affiliated with ESM-Institute any more
}

\date{
	\today
}

\hypersetup{
	pdftitle={Comparison of synthetic dataset generation methods for medical intervention rooms using medical clothing detection as an example},
	pdfauthor={P.~Schuelein, M.~Vetter},
	pdfkeywords={medical clothing detection, Domain Randomization}
}

\begin{document}
\maketitle


\begin{abstract}
\textit{The availability of real data from areas with high privacy requirements, such as the medical intervention space, is low and the acquisition legally complex. Therefore, this work presents a way to create a synthetic dataset for the medical context, using medical clothing as an example. The goal is to close the reality gap between the synthetic and real data. For this purpose,  methods of 3D-scanned clothing and designed clothing are compared in a Domain-Randomization and Structured-Domain-Randomization scenario using an Unreal-Engine plugin or Unity. Additionally a Mixed-Reality dataset in front of a greenscreen and a target domain dataset were used. Our experiments show, that Structured-Domain-Randomization of designed clothing together with Mixed-Reality data provide a baseline achieving 72.0\% mAP on a test dataset of the clinical target domain. When additionally using 15\% of available target domain train data, the gap towards 100\% (660 images) target domain train data could be nearly closed 80.05\% mAP (81.95\% mAP). Finally we show that when additionally using 100\% target domain train data the accuracy could be increased to 83.35\% mAP.}
\end{abstract}


\twocolumn 

\section{Introduction}
\label{sec:Introduction}
There are insufficiently large or no datasets for research questions on camera-based AI systems from the field of medical operating rooms. Use cases for AI systems in this context include sterility of health professionals, whether and where certain medical devices are located or action recognition of health professionals. Furthermore, datasets are, if at all, only institution-related and not publicly available due to data protection regulations and ethics requirements.

The successes of deep learning in recent years are among others, due to the availability of large datasets such as Imagenet \cite{ILSVRC15} for Image Classification, or MS COCO \cite{DBLP:journals/corr/LinMBHPRDZ14} for Object Bounding Box Detection. In addition, the research of new methods or architectures like \cite{He2016}\cite{Lin2020}\cite{Tan_2020}\cite{Bochkovskiy2020} and the research of high performance hardware for parallel computing are to be mentioned. The authors of \cite{Sze2018}\cite{LeCun2019} analyze the hardware topic in depth. In this work however, special focus is put on the availability of datasets and methods for dataset generation in order to reduce the necessary amount of real data from the target domain. 

Several works already deal with the generation of synthetic data and with the goal of reducing the reality gap between synthetic and real data. Among these are the work on Domain Randomization (DR) and Structured Domain Randomization (SDR) \cite{DrTobin}\cite{DrTremblay}\cite{Prakash_2019}. In addition to other work, it has already been shown that the use of synthetic image data can decrease the required amount of real data \cite{DrTremblay}. Likewise, synthetic data of persons exist \cite{h36m_pami}. However, the challenge lies in the specific characteristics of a medical intervention space. Health professionals wear special clothing with sometimes multiple layers, wear sterile gloves, masks and hairnets. The differences between conventional human data and data from the medical field are large. Nevertheless, Domain Randomization techniques sound promising for use in research questions around medical interventions.

This work presents a comparison in terms of detection accuracy and generalizability of different methods for synthetic clothing generation using either 3D clothing scans (SCANS) or designed CAD clothing (CAD) with SMPL models \cite{SMPL:2015}. The comparison is performed using the example of medical clothing object detection. Both methods (SCANS, CAD) are incorporated into a domain randomization environment called NDDS \cite{to2018ndds} and a Structured Domain Randomization environment implemented in Unity, based on \cite{Technologies2017}, to generate synthetic training data. Likewise, the aim of the presented methodology is to explore a pipeline for the generation of synthetic data for the medical field, so that further research questions from the intervention space can be explored. In addition to synthetic data, real data of different persons are recorded in front of a green screen and in a clinical environment. All data are split into train/val/test sets while the clinical dataset serves as a test set for all methods.

\section{Related Work}
\label{sec:relatedWork}
With the rise of synthetic data generation methods, for example Domain Randomization \cite{DrTobin}, it has already been shown that synthetic data can reduce the amount of real data required \cite{DrTremblay}. However, one focus of research is the reduction of the reality gap between the synthetic data and the target domain.

Here, the aforementioned Domain Randomization (DR) has turned out to be one way to reduce the gap. One idea of DR is that if enough variance can be generated in the synthetic data, reality represents another variance of the target domain \cite{DrTobin}.

The work of \textit{Tobin et al.} \cite{DrTobin} and \textit{Tremblay et al.} \cite{DrTremblay} showed, that an object detection network for robot grasping or car detection can be trained from synthetic images with random positioning, random lighting, random backgrounds, distractor objects and non-realisitc textures alone. In addition, the work of \textit{Tremblay et al.} showed, that the necessary amount of real target domain data can be reduced while maintaining adequate accuracy, when pretrained with DR generated images.

Also the work of \textit{Borkman et al.}\cite{Borkman2021} showed that when using Unity Perception for synthetic data generation, the amount of real-world data could be reduced to 10\% when used together with the synthetic data, while achieving better AP score as with all real-world data alone. 

Domain Randomization has already been successfully applied in various fields. In addition to the mentioned areas of car detection and robot grasping, the work of Sadeghi et al. \cite{Sadeghi2017} for flying a quadrocopter through indoor environments, \textit{Zhang et al.} \cite{DBLP:journals/corr/abs-1709-05746} for a table-top object reaching task through clutter or \textit{James et al.} \cite{DBLP:journals/corr/JamesDJ17} for grasping a cube and placing it in a basket can be named.

This leads us to believe that DR is a suitable approach for the medical intervention room domain, where no real data is largely available and access to that domain is widely restricted.

Ablation studies of \cite{DrTremblay} and \cite{DrTobin} showed, that high resolution textures and higher numbers of unique textures in the scene improve performance.
Also, \cite{DBLP:journals/corr/abs-1709-05746} come to the conclusion, after testing their hypothesis, that using complex textures yields better performance than using random colors.

In contrast to the domain randomization approach is the photorealistic rendering of the scene and objects. A number of datasets have been created for this purpose in recent years. Here the works of \cite{Tremblay2018a} \cite{h36m_pami} \cite{Varol2017LearningFS} or \cite{GraspingDRphoto} are to be mentioned. Some of these works combine real image data with domain randomization and photorealistic rendered image data.

In \cite{Tremblay2018a} a photorealistic rendered dataset was created for 21 objects of the YCB dataset. Here, the objects are rendered in different scenes with collision properties when falling down. The dataset is intended to accelerate progress in the area of object detection and pose estimation. 

In \cite{GraspingDRphoto}, domain randomization is combined with photo realistically rendered image data, for robotic grasping of household objects. Using the data generated in this way, the authors have managed to explore a real-time system for object detection and robot grasping with sufficient accuracy. They also showed that the combination of both domains improved performance as opposed to just one alone.

In the field of human pose estimation, the works of \cite{h36m_pami} and \cite{Varol2017LearningFS} need to be mentioned. Both works were able to show that the performance of networks can be increased by using synthetic and animated persons, respectively.

The work of \cite{Varol2017LearningFS} generates photo-realistic synthetic image data and their ground truth for body part classifications.

In \cite{h36m_pami}, animated persons are integrated into mixed reality environments. The movements were recorded by actors in a motion capture scenario and transferred to 3D scanned meshes. In their experiments, they were able to achieve a 20\% increase in performance compared to the largest training set available in this domain.

State of the art models for realistic human body shapes are the SMPL models introduced by \cite{SMPL:2015} and improved by STAR in \cite{Osman2020}. SMPL stands for Skinned Multi-Person Liner Model (SMPL) which according to the authors is a skinned vertex-based model which represents human shapes in a wide variety. In their work they learn male and female body shape from the CAESAR dataset \cite{CAESAR}. Their model is compatible with a wide variety of rendering engines like Unity or Unreal and therefore highly suited to be used in synthetic data generation for humans. There also exist extensions to the SMPL model like MANO and SMPL-H which introduce a deformable hand model into the framework. MANO \cite{MANO:SIGGRAPHASIA:2017} is learned from 1000 high resolution 3D scans of various hand poses.

\section{Method}
\label{sec:methods}
As previously mentioned, real-world data collection in medical intervention rooms is complex, costly, and requires approval from an ethics board and the persons involved. 
As shown in the previous, DR can help train an object detection network with sufficient performance in real world applications.
 
However, one challenge in dataset generation for the medical intervention space is domain-specific clothing. We argue, that randomizing the clothing textures with random textures would help improve detection rates of the clothing types, but when applied in real world applications, for example a colored T-shirt would not be distinguishable from the targeted blue colored specific area clothing. For the general detection of cars as in \cite{DrTremblay} the randomization technique makes sense, but for the domain specific use case presented here something else should be used in our opinion.


The questions we try to address in this work are:
\begin{enumerate}
\item How can health professionals be modeled for synthetic data generation?
\item Which techniques are best suited for SDR/DR clothing generation?
\item Can we close the reality gap further by including greenscreen data (Mixed Reality, MR)?
\item Can the required amount of real data be reduced by using SDR/DR/MR ?
\end{enumerate}

For point 1) we argue to use a deformable human shape model like the SMPL models. This provides sufficient variance of different human shapes and sizes. For point 2) we explore two different methods of clothing generation. First we 3D scan various persons wearing medical clothing and generate a database of different medical clothing scans for each clothing type, which we call SCANS. Second we commission a professional graphics designer to create assets based on the area clothing, which we call CAD. Regarding point 3) we take images in front of a greenscreen of different persons wearing medical clothing which we label by hand. For point 4) we explore wether and by how much the required amount of real data can be reduced.

To address the named questions further, we set up experiments where we want to detect the following classes with the help of the Scaled Yolov4 object detector \cite{Wang_2021_CVPR}

The classes to be detected are:
\begin{itemize}
	\item humans
	\item area clothing shirt
	\item area clothing pants
	\item sterile gown
	\item medical face mask
	\item medical hairnet
	\item medical gloves
\end{itemize}

\begin{figure}
	\centering
	\begin{tabular}{l c}
		\includegraphics[width=.22\textwidth]{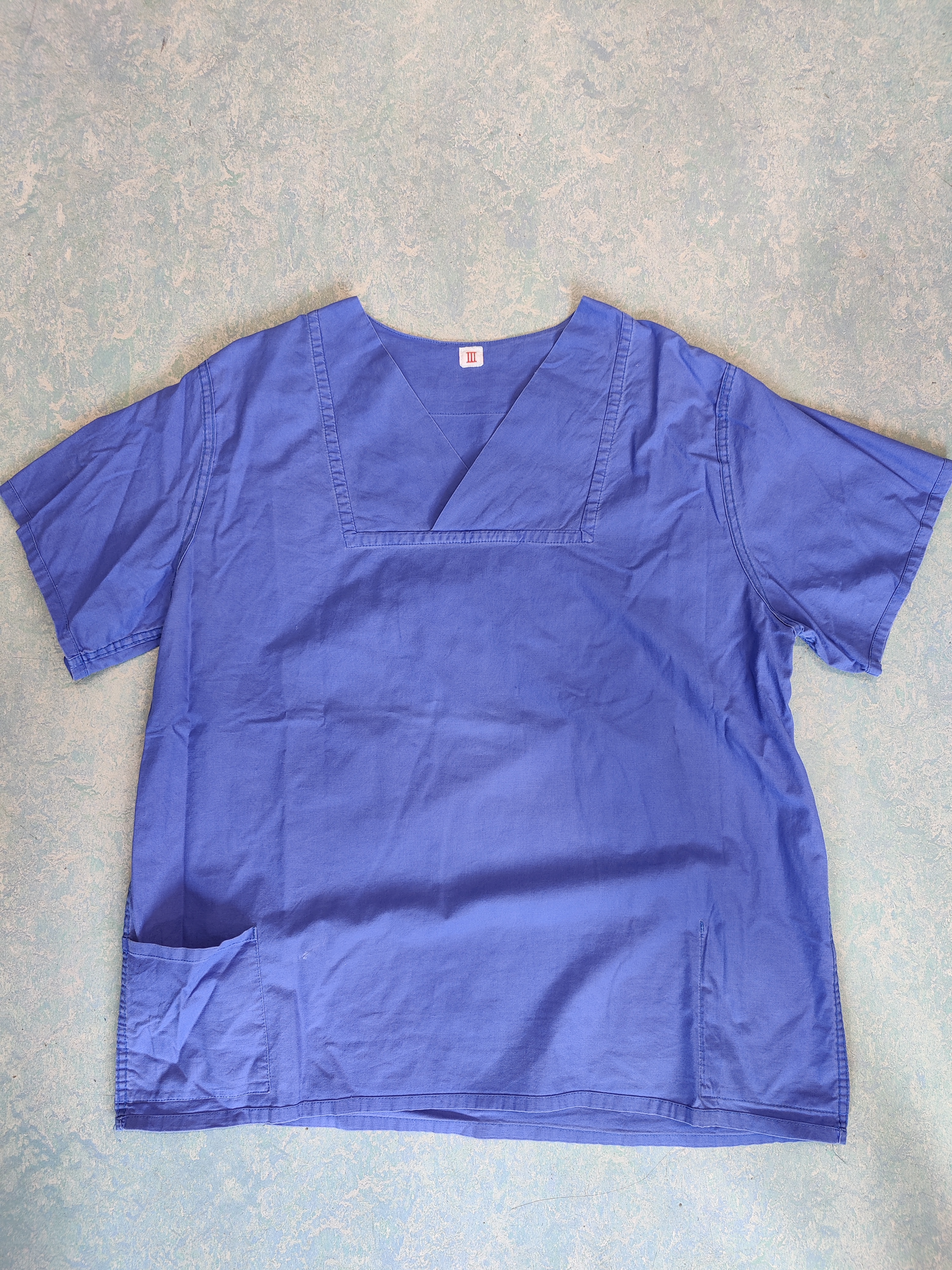}
		&
		\includegraphics[width=.22\textwidth]{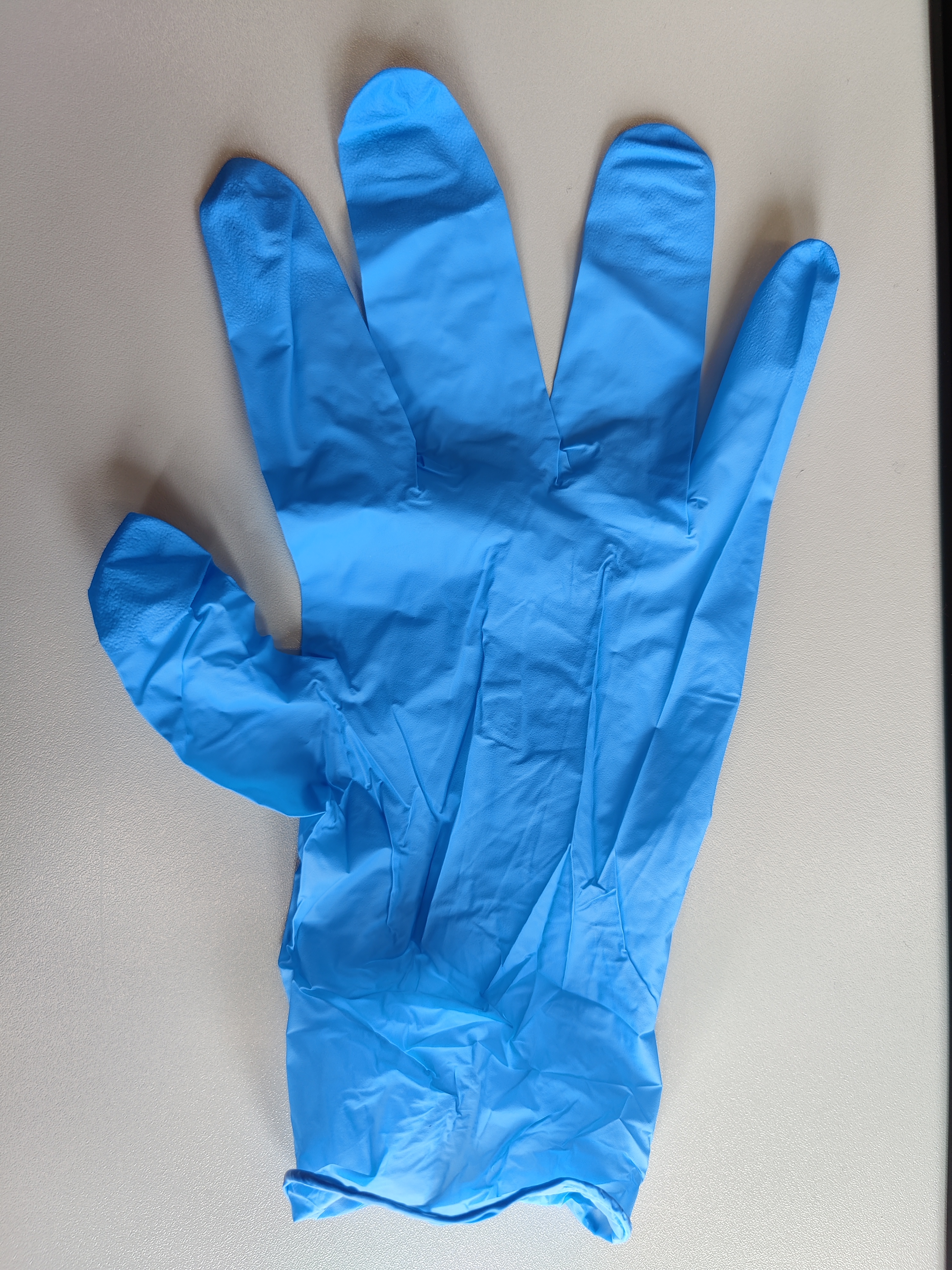} \\
		
		\includegraphics[width=.22\textwidth]{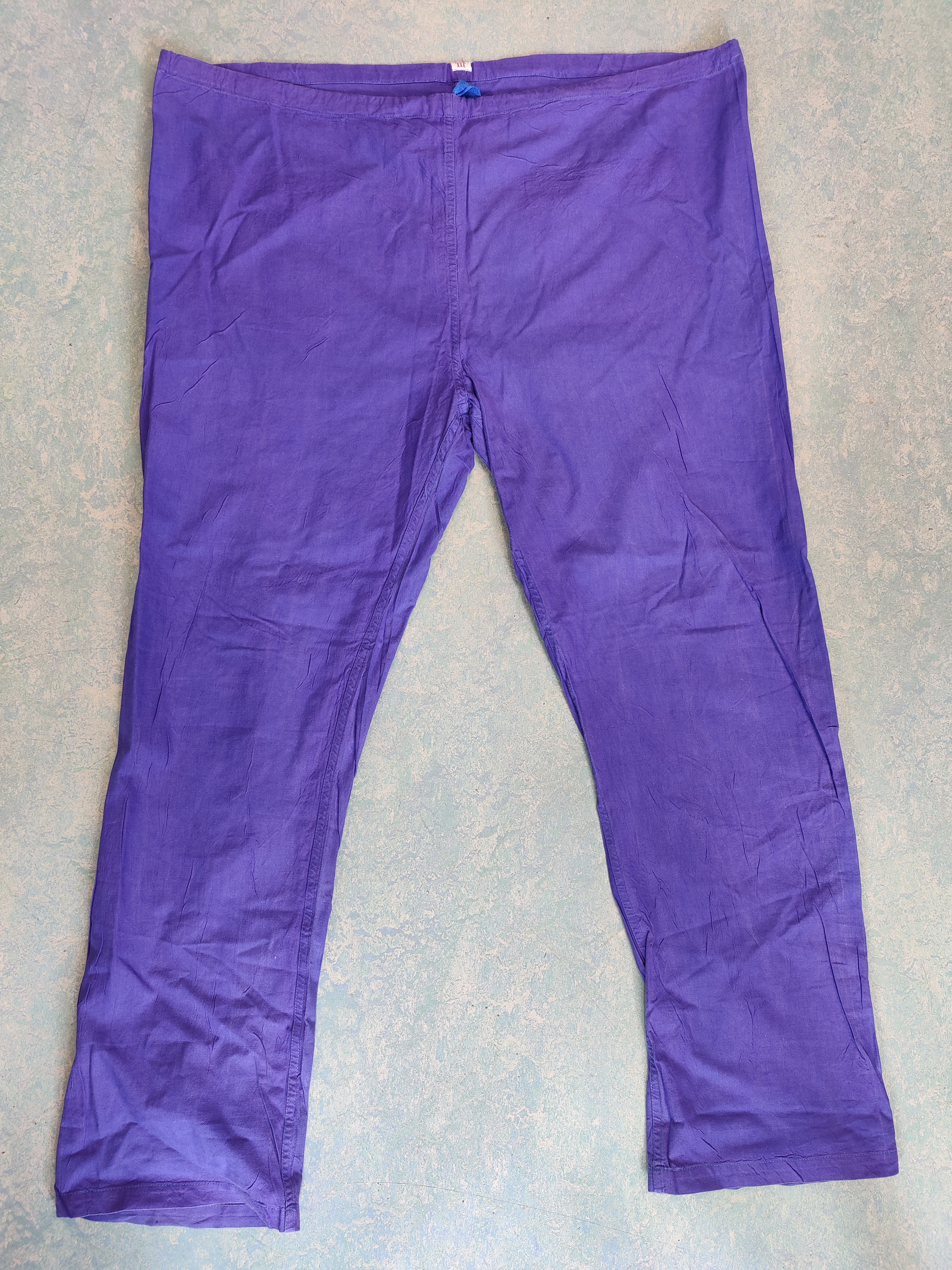}
		&
		\includegraphics[width=.22\textwidth]{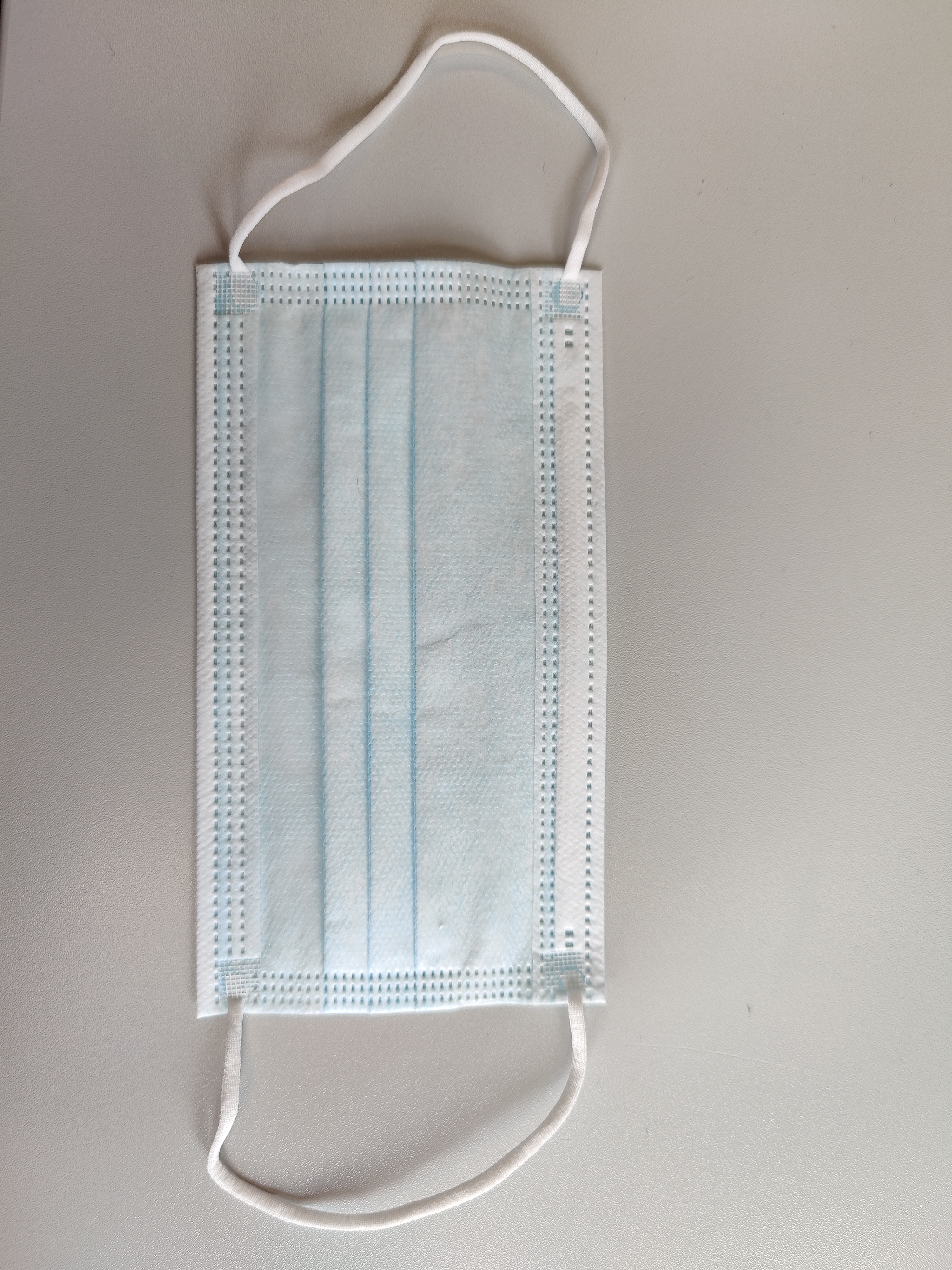} \\
	\end{tabular}	
	\caption{Medical clothing examples like area shirt, pants, mask and glove. These clothing types among others represent the target clothing for our object detection network.}
	\label{fig:expClothing}
\end{figure}

Examples of the medical clothing are given in figure \ref{fig:expClothing}.

\subsection{Character Creation}
\label{sec:charCreation}
The medical characters we use in SDR/DR are built through a combination of SMPL body models, textures, animations and clothing assets. Within the following sections each of the components are described in detail. 

\textbf{Body Model}\\
As the base of our characters we use the male and female model of the SMPL+H model from \cite{MANO:SIGGRAPHASIA:2017}. The models cover a huge variety of realistic human shapes, which can be randomized through ten blend shapes. We decide to use the extended SMPL+H model instead of the original SMPL model \cite{SMPL:2015}. This is because one of our cloth items are gloves and through the hand rig of the SMPL+H model we will be able to create more deformations of the glove asset. 

\textbf{Human Texture}\\
To add more variation and realism to the appearance of the characters, the texture maps from \cite{Varol2017LearningFS} are used.
Out of the 930 textures, only 138 (69 of every gender) have been used. This is, as we created our own cloth assets, only the textures of people in undergarments were relevant. Those texture maps were created out of 3D body scans from the CAESAR dataset \cite{CAESAR} and cover a variety of skin colors and identities, however all of the faces have been anonymized \cite{Varol2017LearningFS}. 

\textbf{Pose} \\
To provide a variety of realistic body poses, the models were animated through Motion Capture (MoCap) data, which has been captured within our laboratory. We track the movement of 74 joints down to the fingertips. We use an intrinsic Motion Capture suite with the Hand gloves Add-on called Perception Neuron Studio \footnote{Perception Neuron Studio suite and gloves addon: https://neuronmocap.com/perception-neuron-studio-system}. In order to keep the dataset simple, we only used one animation in our experiments. The potential to add more varying animations is given however.\\

\begin{figure}
	\centering
	\begin{tabular}{l c}
		\includegraphics[width=.22\textwidth]{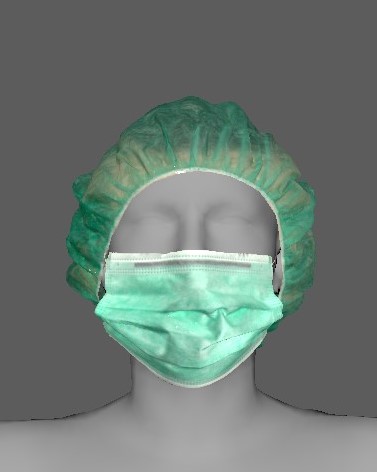}
		&
		\includegraphics[width=.22\textwidth]{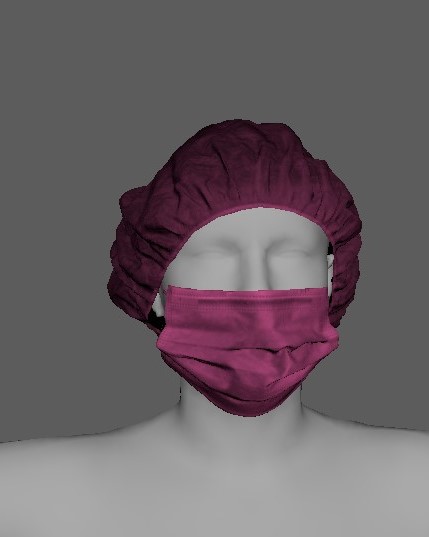} \\
		
		\includegraphics[width=.22\textwidth]{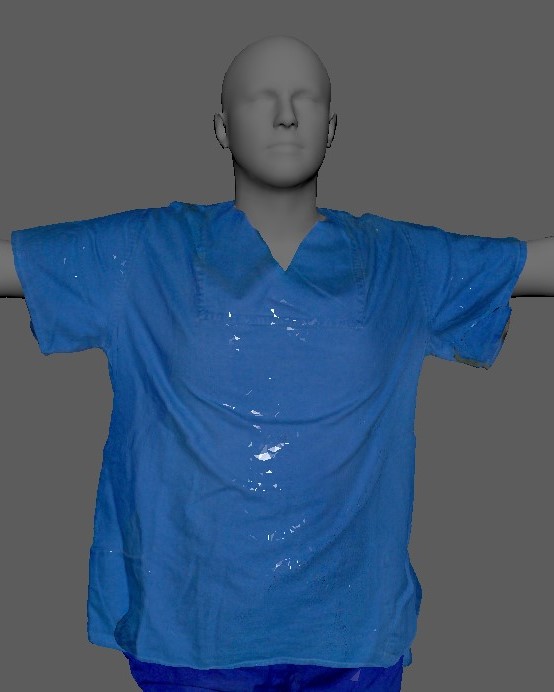}
		&
		\includegraphics[width=.22\textwidth]{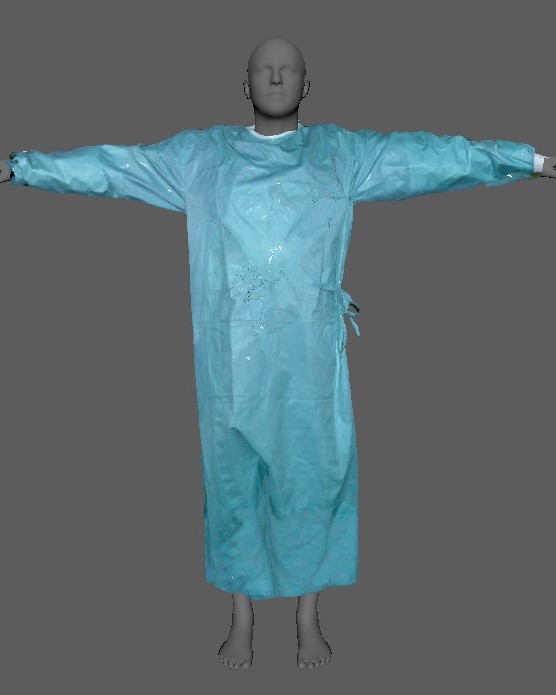} \\
	\end{tabular}	
	\caption{Examples of our 3D-scanned clothing assets with color augmentation.}
	\label{fig:expClothingScans}
\end{figure}

\textbf{3D-scanned Cloth Assets (SCANS)} \\
A 3D Scanner called Artec Leo\footnote{3D Scanner Artec Leo: https://www.artec3d.com/portable-3d-scanners/artec-leo } with a 3D resolution of 0.2mm was used to capture the medical cloths. For our synthetic training dataset we used clothing scans of 4 male and 4 female models. In this way, variations of the real world textures, including reflections, wrinkles, colors and surface texture information are collected. After building an initial model from the 3D Scanner, we adapt the cloths to fit the standard male and female SMPL+H Character using 3D Modelling techniques. As the gloves should match the Characters fingers exactly, to ensure correct animations of the fingers, they have been modeled instead of scanned. Afterwards the textures of scanned gloves have been applied to the models. When the cloth assets are fitted, we create cloth blend shapes, which match the blend shapes of the SMPL+H model, in order to make them adaptable with the character. Additionally, the cloth meshes are bound to the same rig as the character, by transferring the skin weights of the SMPL+H models. Like this the cloth assets are just as adaptable in shape and pose as the body model. 
According to our research, medical cloths usually come in the colors blue, green and light pink. To cover this variation in our dataset, without scanning more cloth or performing augmentations on the whole image, we augmented the texture maps. Examples of the scanned and rigged clothing assets can be seen in figure \ref{fig:expClothingScans}.

\begin{figure}
	\centering
	\begin{tabular}{l c}
		\includegraphics[width=.22\textwidth]{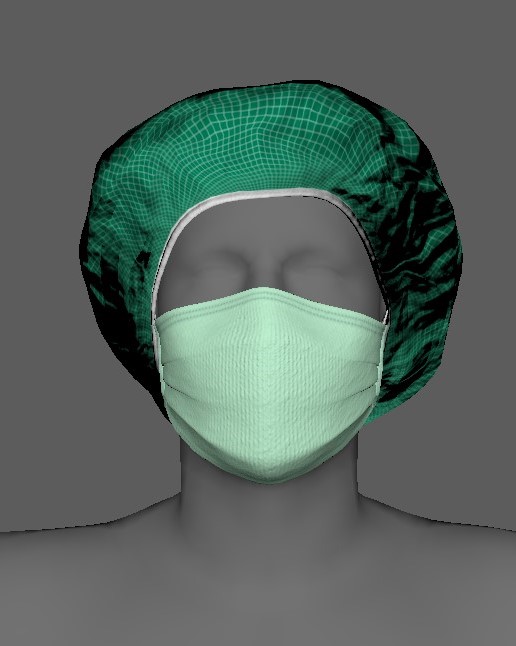}
		&
		\includegraphics[width=.22\textwidth]{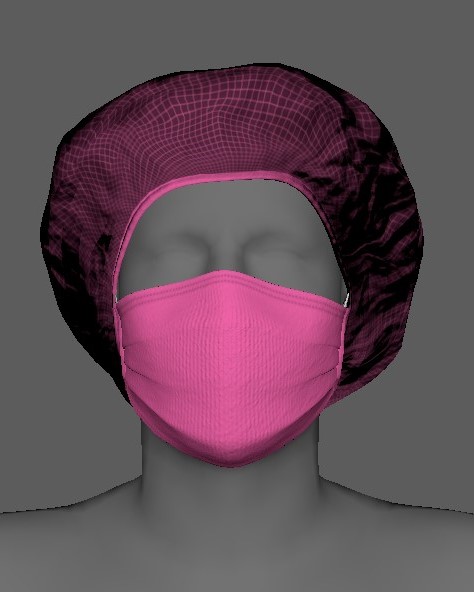} \\
		
		\includegraphics[width=.22\textwidth]{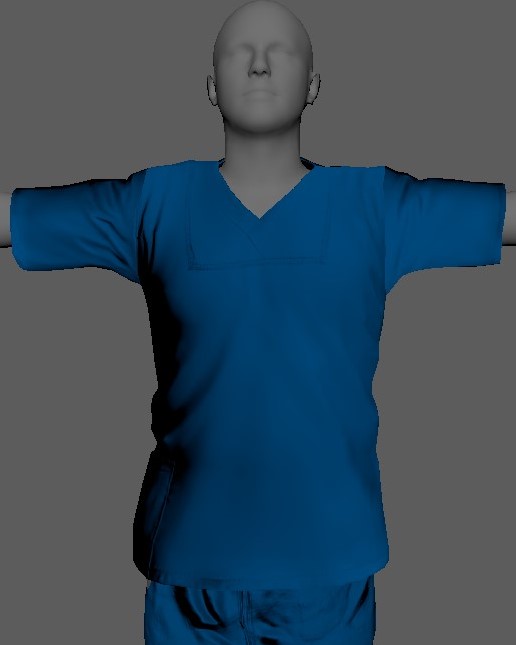}
		&
		\includegraphics[width=.22\textwidth]{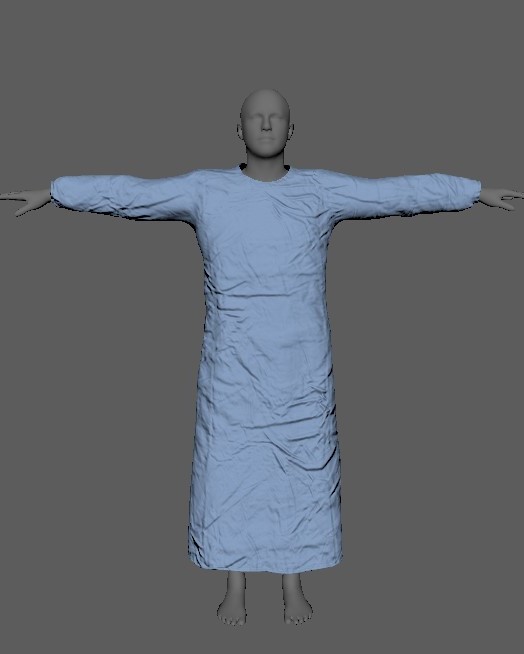} \\
	\end{tabular}	
	\caption{Examples of the designed clothing assets with color augmentation.}
	\label{fig:expClothingCad}
\end{figure}

\textbf{Designed Cloth Assets (CAD)} \\
To evaluate the performance of 3D-scanned cloth assets we compare them to hand designed clothing assets. Therefore we have asked a designer\footnote{azeemdesigns: https://www.fiverr.com/azeemdesigns?\\source=order\_page\_user\_message\_link} on Fiver to model the clothes. Examples of those assets can be seen in figure \ref{fig:expClothingCad}. We first evaluated to what extent freely available assets from the assets stores can be used for this purpose. However, there are no assets available that match our specific clothing in total. Therefore, we have decided to have the assets designed. The designed assets have been processed in the same way as our scanned assets. They are also deformable and are bound to the same rig.

\begin{figure}
	\centering	
	\includegraphics[width=.49\textwidth]{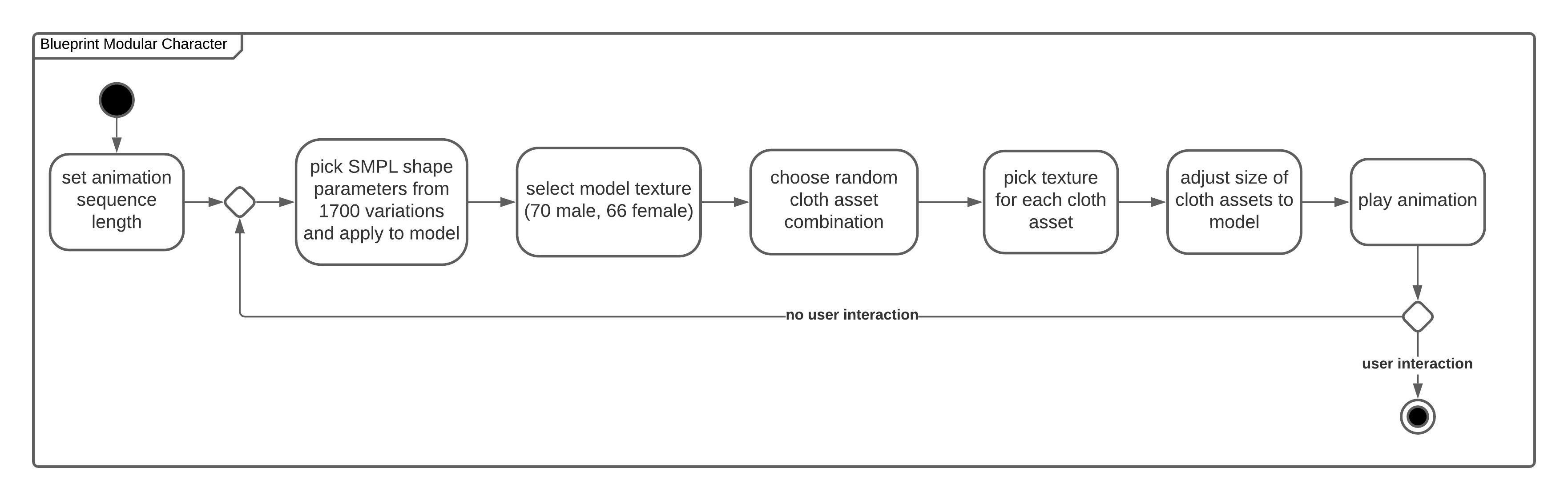} \\	
	\caption{Activity Diagram of the modular character blueprint used in Unreal with NDDS.}
	\label{fig:flowChartBlueprint}
\end{figure}

\textbf{Modular Character DR, NDDS} \\
To create the datasets from the body models, animations, cloth assets and all other components are combined in a character blueprint in Unreal Engine 4.22.3(DR). Here we have created a modular character, who is able to take on random shapes, textures and combinations of clothing. Also the 3D scanned cloths and designed cloths will vary in shape with the character and move with the animation. When the data capturing begins, the character first iterates over 1700 body shapes. The SMPL shape parameters for these body shapes have been taken from \cite{Varol2017LearningFS} and represent 1700 male and female body types taken from the CEASAR Dataset. Next, one randomly sampled human texture is applied to the body model. Afterwards one cloth item out of each category is randomly sampled, textured and added to the body model. After the blend shapes of all cloth assets are adapted to the current body shape, one animation is chosen. When the animation finished playing, the next body shape will be chosen and the process will repeat itself. We create two seperate datasets, one with 3D-scanned (SCANS) assets and another with designed (CAD) assets for DR. An activity diagram, which represents the blueprint for Modular Character creation in NDDS, is given in figure \ref{fig:flowChartBlueprint}.\\

\begin{figure}
	\centering	
	\includegraphics[width=.49\textwidth]{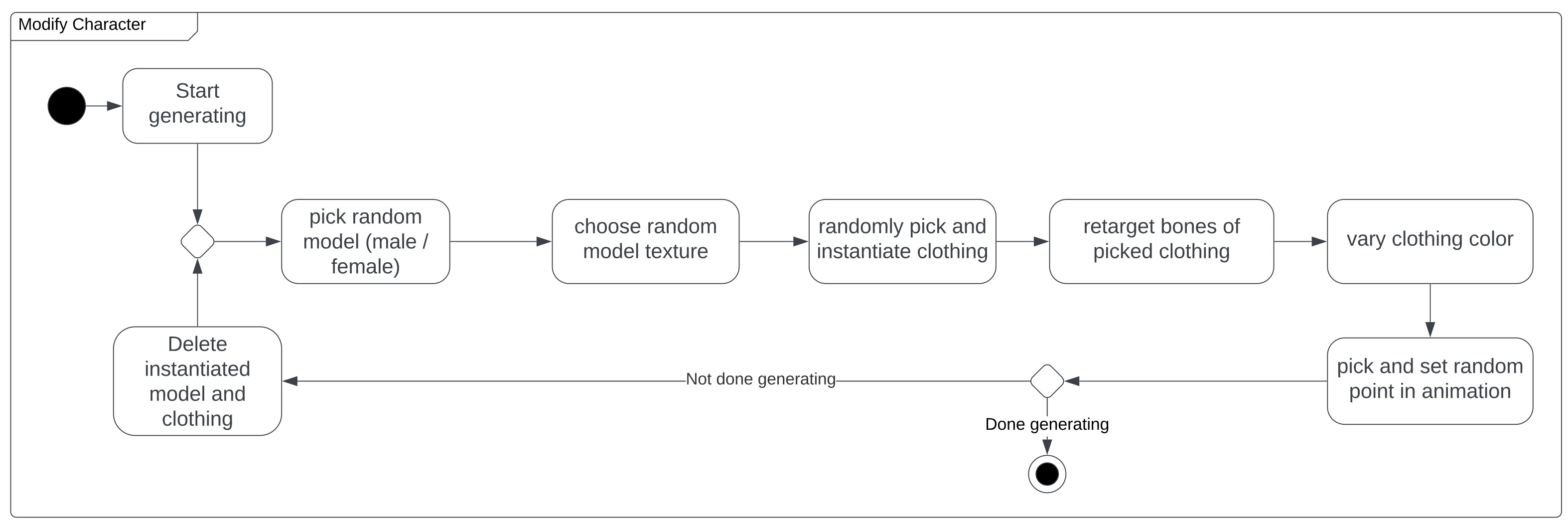} \\	
	\caption{Activity Diagram of the character creation used in Unity for SDR.}
	\label{fig:flowChartUnity}
\end{figure}

\textbf{Modular Character SDR, Unity}\\
To be able to use SMPL models in unity, the male and female models are converted to a Humanoid character and an avatar is created from the model. Variation of clothing in unity is made possible using a custom-made component which selects random pieces of clothing from each category and applies them to a randomly selected SMPL model. In order to enable the clothing to animate together with the model, the bones of each piece of clothing are changed to match with those of the model using said component.  Afterwards a random animation can be assigned to the model using the Animator component with a custom Animator Controller by setting the Motion parameter. Finally the model including all selected clothing is instantiated and randomly placed inside the scene. The animation is then moved along during data generation by setting the Motion Time parameter of the Animator Controller to a random value. The character itself is adapted using the custom-made randomization components by varying the texture of the models as well as the material of the clothing using predefined color variations. Models as well as clothing and position are varied each frame. The shape of the models is not modified as it resulted in clipping of the clothing and other unrealistic behavior. An activity diagram is shown in figure \ref{fig:flowChartUnity}

\begin{figure}
	\centering
	\begin{tabular}{l c}
		\includegraphics[width=.23\textwidth, trim={5cm 2cm 25cm 2cm},clip]{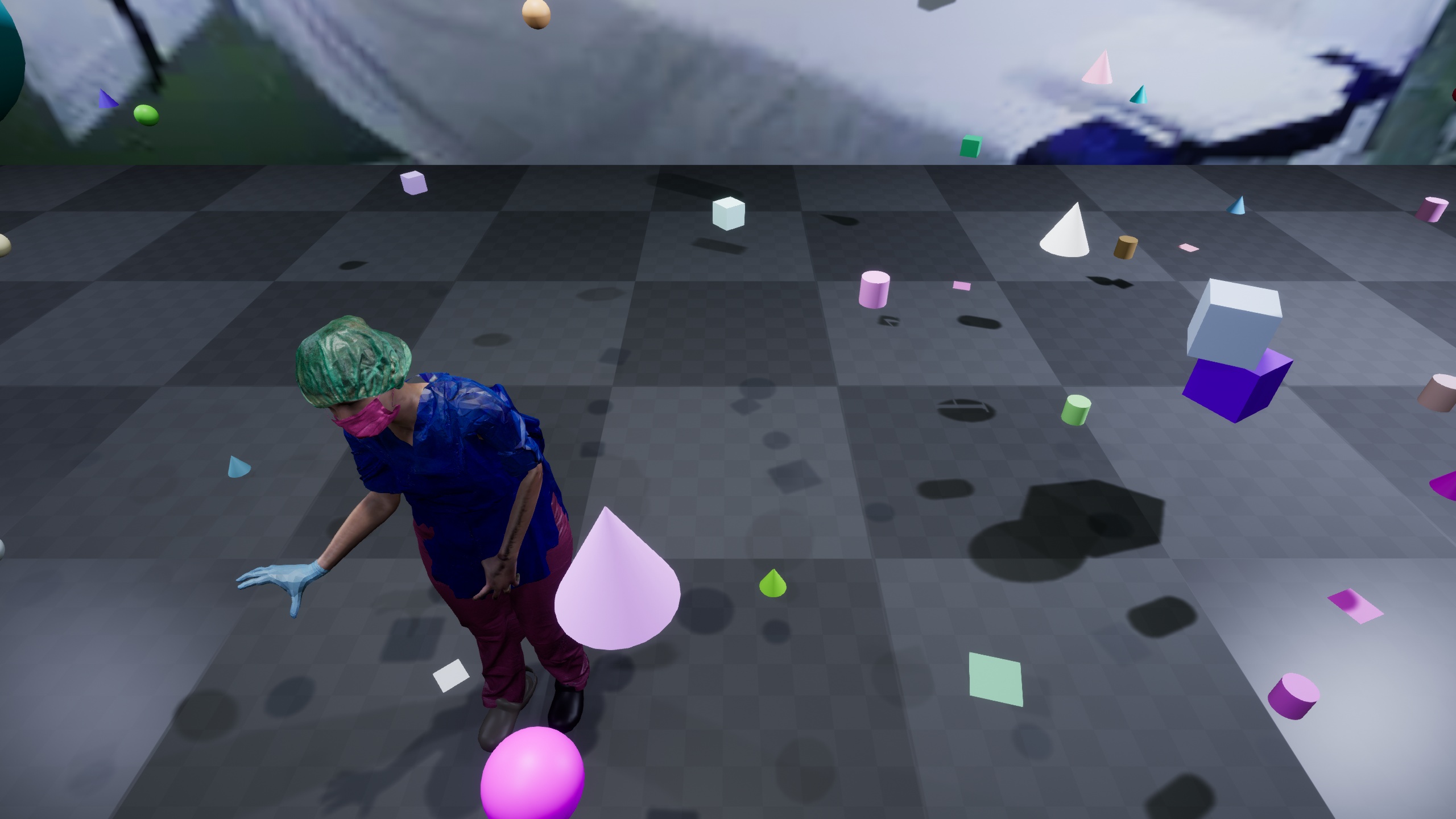}
		&
		\includegraphics[width=.23\textwidth, trim={25cm 2cm 15cm 9.5cm},clip]{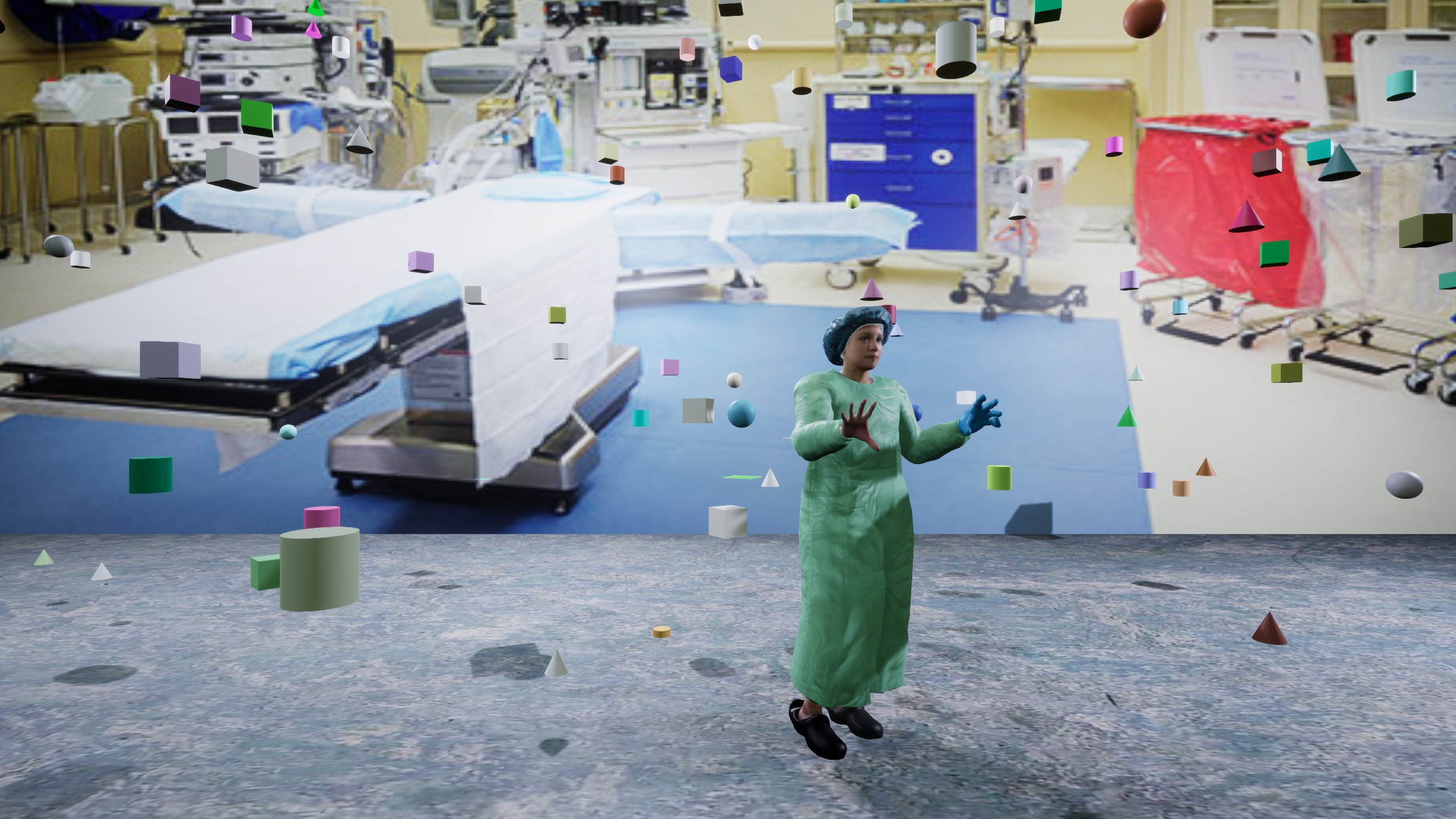} \\
		
		\includegraphics[width=.23\textwidth, trim={25cm 2cm 2.5cm 9.5cm},clip]{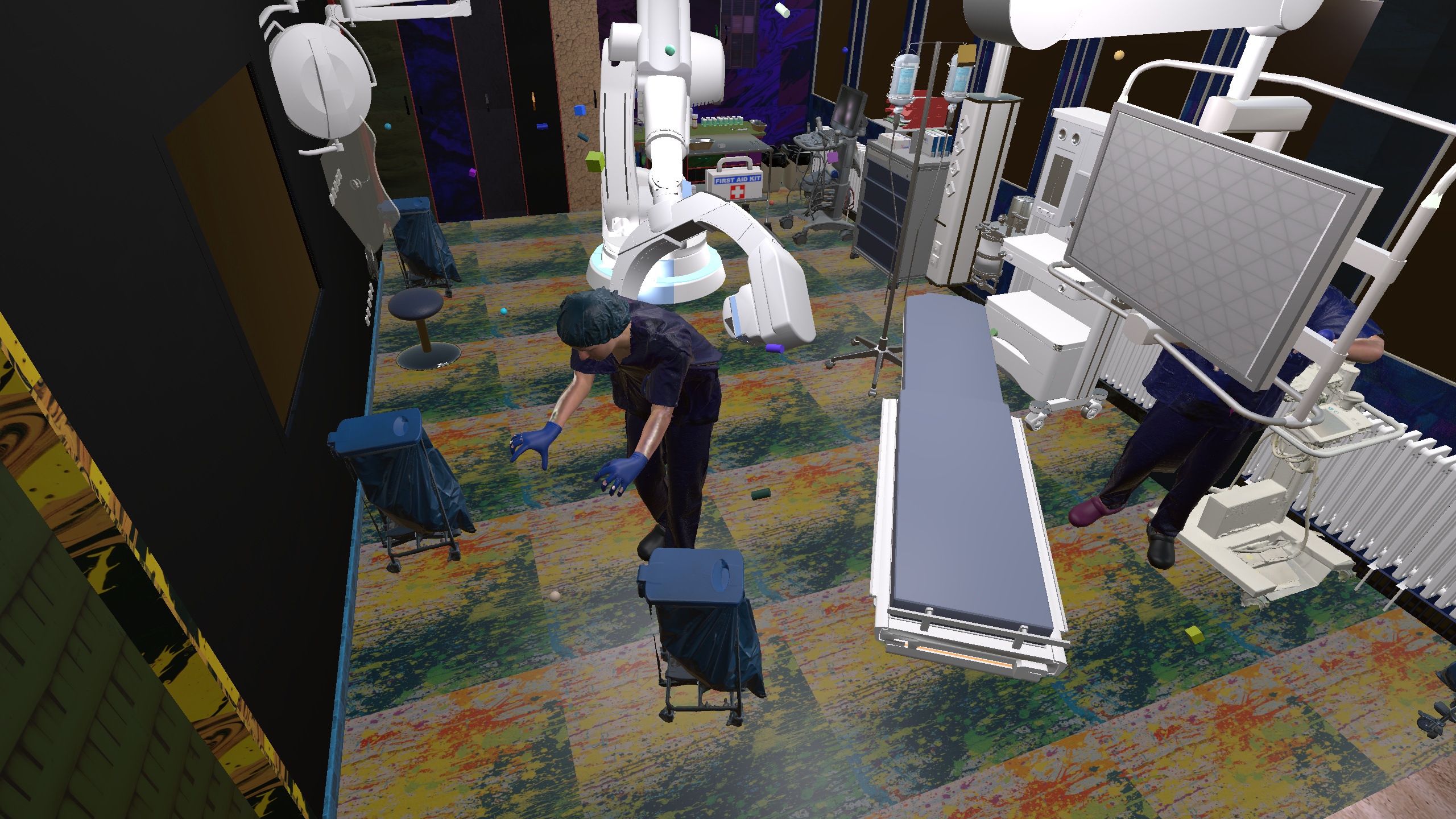}
		&
		\includegraphics[width=.23\textwidth, trim={25cm 2cm 2.5cm 4.5cm},clip]{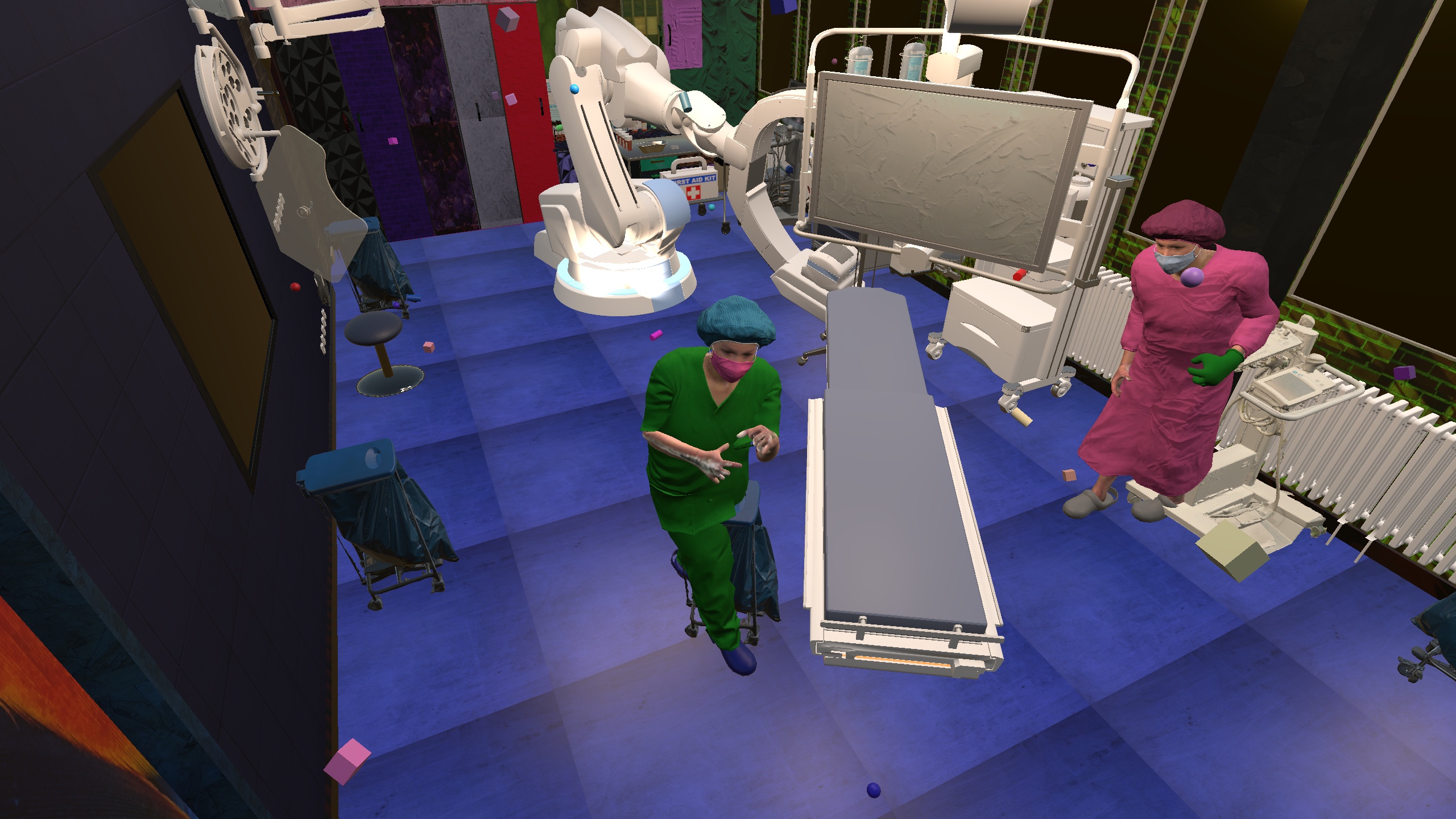} \\
	\end{tabular}	
	\caption{Examples of synthetic RGB image data from DR and SDR datasets (top: DR, bottom: SDR, left: SCANS, right: CAD).}
	\label{fig:expSynth}
\end{figure}

\textbf{Domain Randomization}
For the synthetic data generation of DR, an Unreal Engine 4 plugin called NDDS\cite{to2018ndds} was used. This allows the generation of RGB images at rates similar to real cameras, as well as depth image data and segementation masks of the scene within Unreal Engine 4 (UE4). The plugin also creates bounding box labeling data for each object in the scene in 2D and 3D. The tool was specifically developed for Domain Randomization and therefore provides tools for scene randomization like object or camera position, lighting and distractor objects, among others. Using the aforementioned Modular Character blueprint, NDDS enables the generation of synthetic datasets for sterile clothing using 3D scanned clothing or designed based clothing. Example images are given in figure \ref{fig:expSynth} on the top row.

\textbf{Structured Domain Randomization}
For dataset generation of SDR, we used a Unity plugin called ML-ImageSynthesis \cite{Technologies2017}  as a base and adapted it to work with the universal rendering pipeline (URP) for quality improvement. Using Unity 2020.3.32f1, additional components have been added to enable an export of additional metadata regarding each generated image such as camera parameters, bounding boxes and world position in .json format. SDR is made possible by making use of a variety of custom-made components which allow the randomization of parameters such as lighting, material, texture, position. The plugin ProBuilder provided by Unity was used to build an intervention room based on the target domain of the real dataset (Klinikum). Scene randomization is achieved by utilizing the aforementioned randomization components.

\subsection{Datasets}
\label{sec:dataset}
To investigate the potential accuracy difference between 3D scanned clothing (SCANS) and designed clothing (CAD) and the amount of necessary real data, different datasets were generated.

First, synthetic datasets of DR and SDR were generated for both scanned and designed clothing, using the presented pipelines in Unreal-Engine und Unity. Second, a dataset in front of a greenscreen was collected (MR-DR). It consists of 8 persons in the training dataset and 2 persons in the validation dataset. The recorded persons move in front of the green screen with a certain grasping motion, which is also used as motion animation for the synthetic data. Finally a dataset of the target domain was recorded (Klinikum). It serves as a baseline comparison for all models and also presents the test data. All datasets are divided into training and validation data. 

\begin{figure}
	\centering
	\begin{tabular}{l c}
		\includegraphics[width=.23\textwidth,trim={15cm 5cm 25cm 2cm},clip]{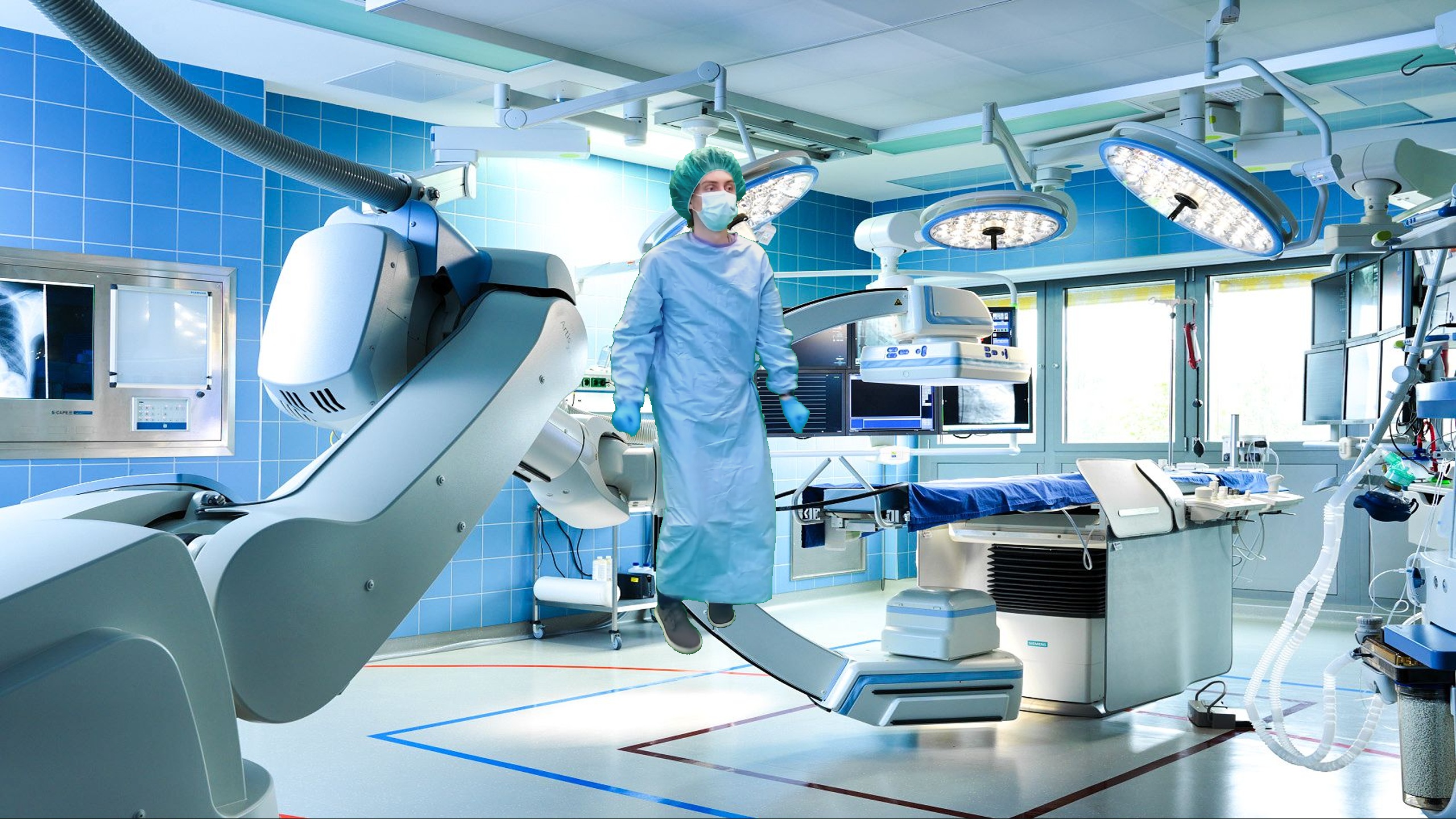}
		&
		\includegraphics[width=.23\textwidth, trim={20cm 2cm 20cm 5cm}, clip]{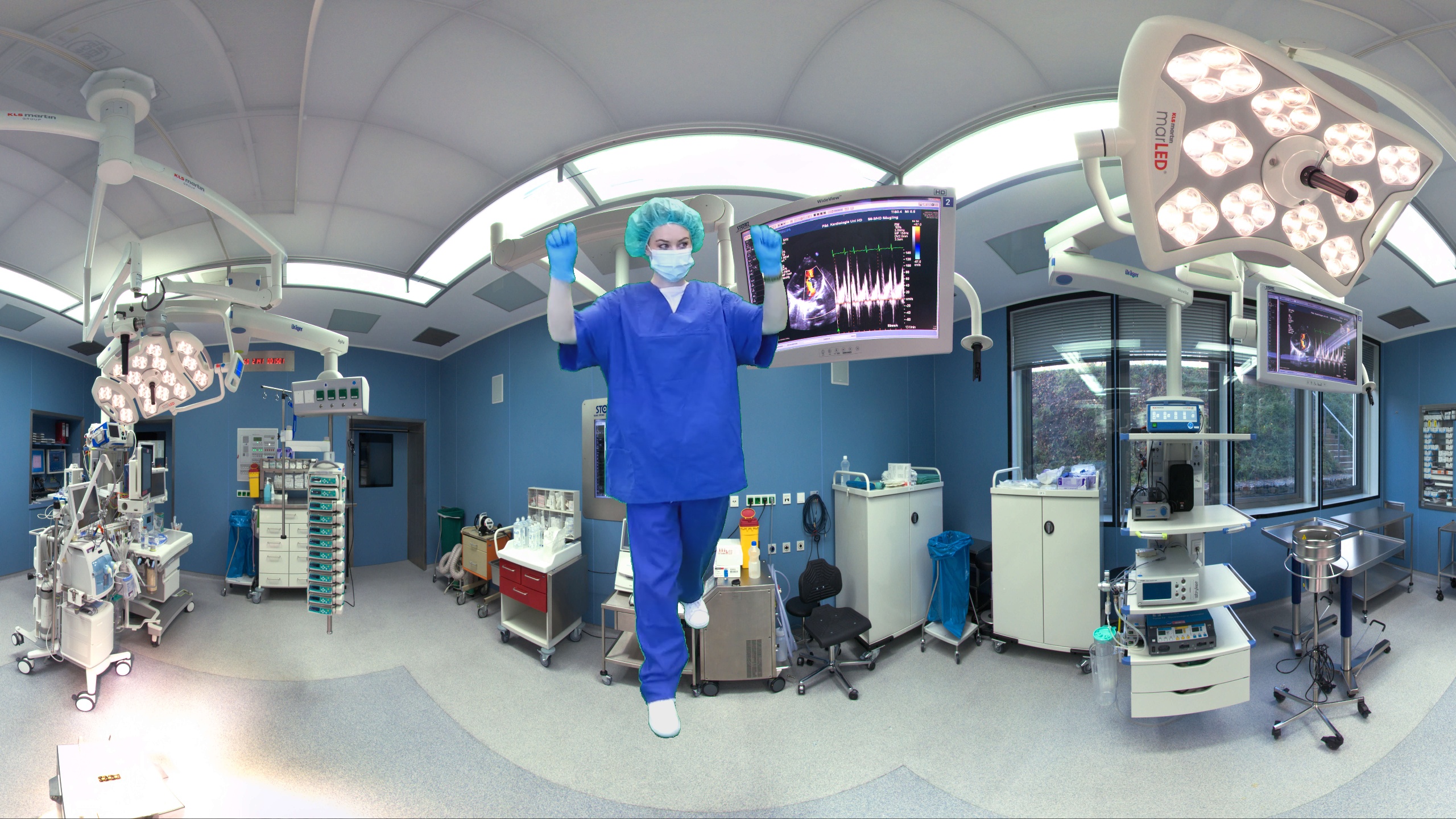}
	\end{tabular}	
	\caption{Examples of the greenscreen dataset with exchanged backgrounds.}
	\label{fig:realDataExps}
\end{figure}

\begin{figure}
	\centering
	\begin{tabular}{l c}
		\includegraphics[width=.23\textwidth,trim={20cm 5cm 20cm 5cm},clip]{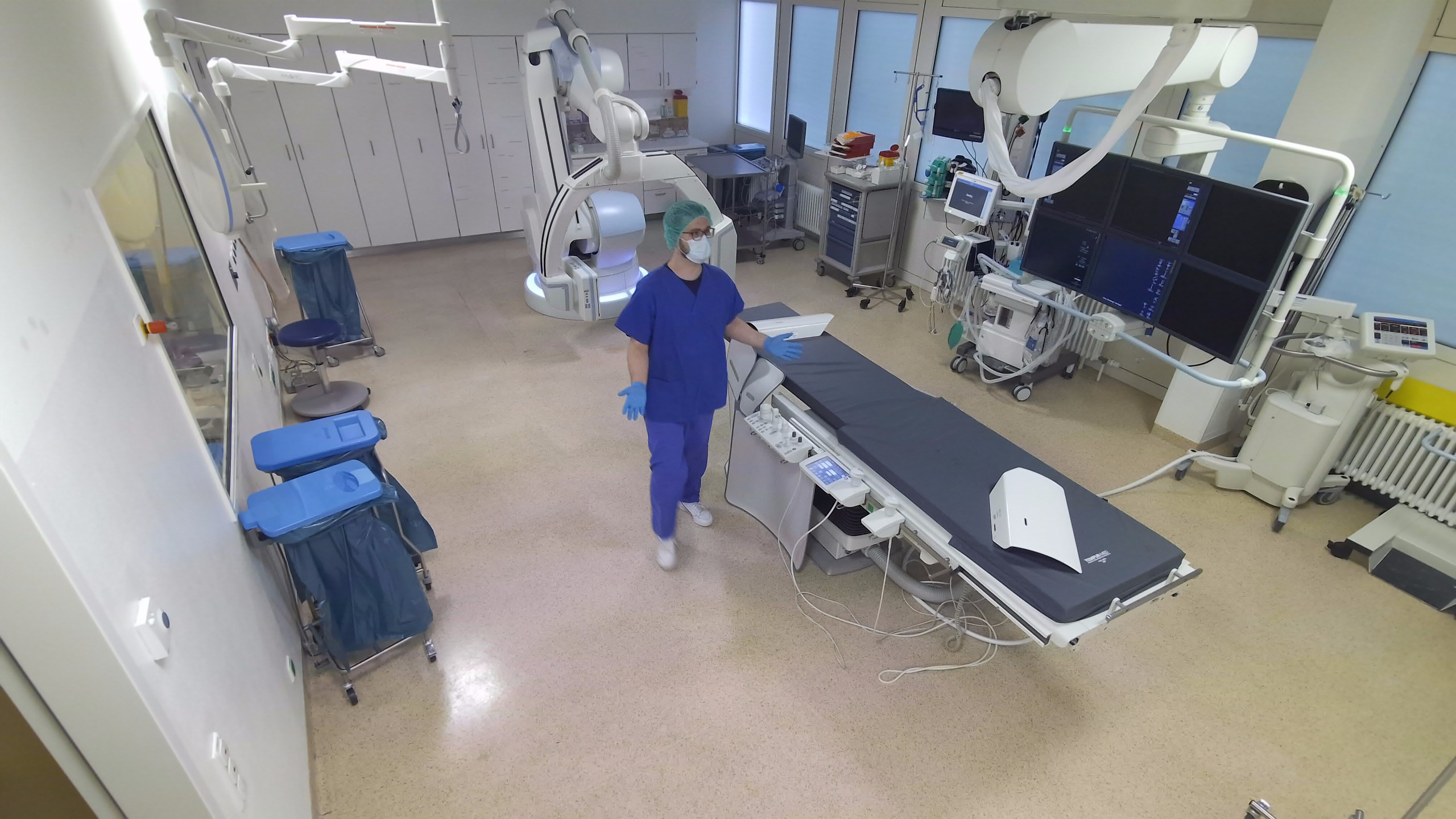}
		&
		\includegraphics[width=.23\textwidth, trim={20cm 5cm 20cm 5cm}, clip]{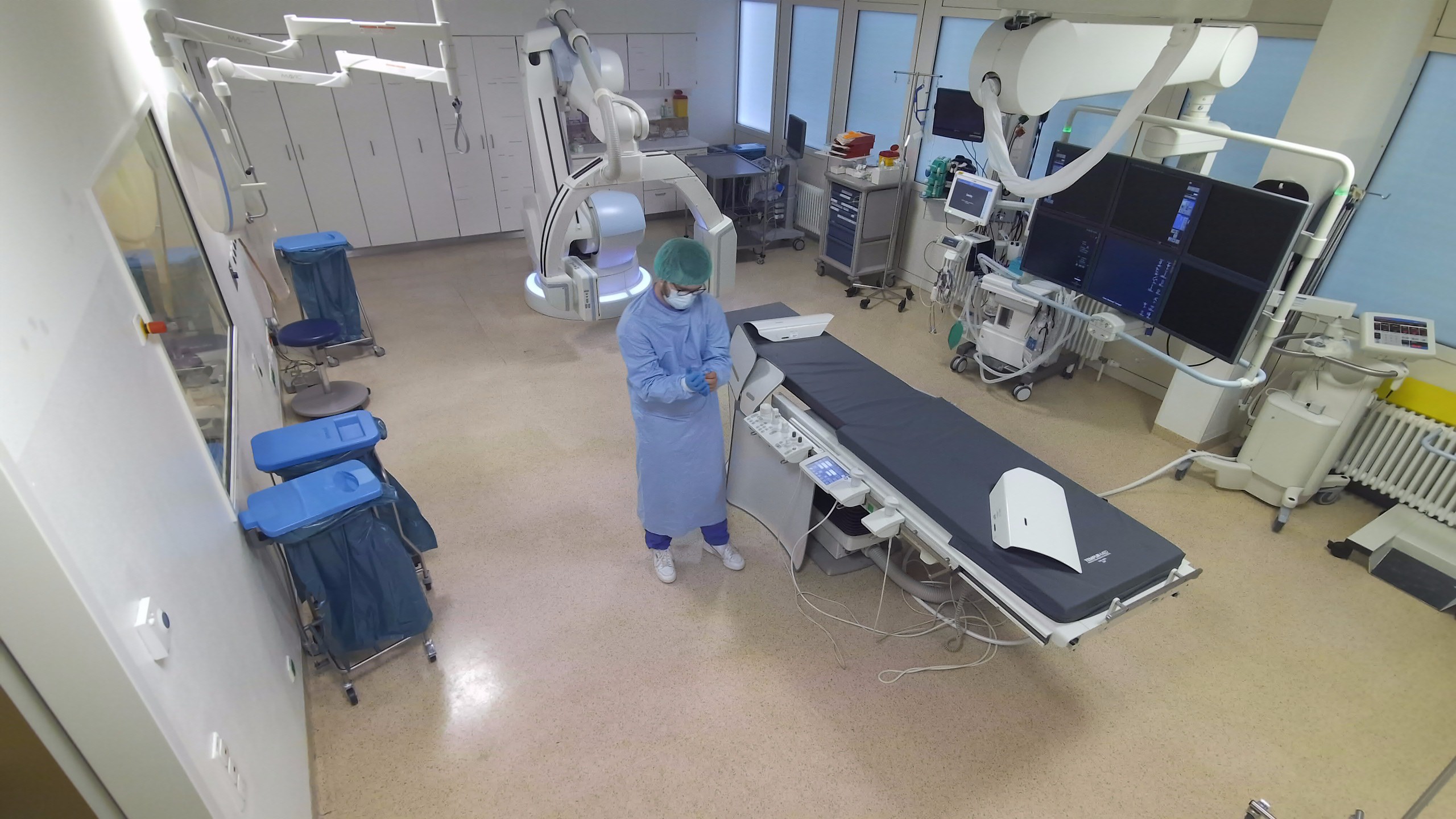}
	\end{tabular}	
	\caption{Examples of the Klinikum test dataset.}
	\label{fig:expKlinikumTest}
\end{figure}

Examples of real data in front of the green screen with exchanged background can be seen in figure \ref{fig:realDataExps}. Examples of the synthetic data can be seen in figure \ref{fig:expSynth} and finally examples from the clinical test data can be seen in figure \ref{fig:expKlinikumTest}.

Table \ref{tab:datasetDistr} shows a breakdown of the sizes and distributions of the datasets.

\begin{table}
	\caption{This table provides data distributions in the different datasets.}
	\centering
	\begin{tabular}{llll}
		\toprule
		\multicolumn{4}{c}{dataset distributions}                   \\
		\midrule
		dataset name 	& train 	& validation 	& test \\
		\midrule
		DRscans 		& 10000		& 1000			& / 	\\
		DRcad  			& 10000		& 1000			& /		\\
		SDRscans		& 8000		& 2000			& /	 	\\
		SDRcad			& 8000		& 2000			& /	 	\\
		MR-DR(GS) 		& 6443		& 1536			& /	\\
		Klinikum		& 660		& 110			& 331	\\
		\bottomrule
	\end{tabular}
	\label{tab:datasetDistr}
\end{table}

\section{Experiment Results}
\label{sec:experiments}

For the experimental investigation of whether and how well 3D scanned clothing compares to designed CAD clothing for detection in the medical environment and how much addition real data is needed to achieve sufficient accuracy, various tests were carried out.

For our experiments we used Scaled-Yolov4 implementation from GitHub\cite{Wang_2021_CVPR}.
At first, 6 different baseline networks were trained to show a basic comparison of the different methods. These baseline models include trainings with synthetic (DRscans, DRcad, SDRscans, SDRcad), mixed-reality (MR-DR) and real data from the clinic domain (Klinikum train).

Training was conducted with default finetune parameters provided from Scaled Yolo-V4\footnote{https://github.com/WongKinYiu/ScaledYOLOv4}. Only Mosaic Augmentation ratio parameters $\alpha$ and $\beta$ were increased from $8.0$ to $20.0$. Additionally a green channel augmentation was used when MR data was present in the training dataset in order to reduce the greenscreen reflection influence which we had troubles with in some classes. We experimentally found out that this gives the best results. All networks were trained for 300 epochs and achieved convergence. Tests were performed with IoU-threshold: 0.5 and confidence-threshold: 0.2. The used Yolov4 network was yolov4-p5, image size setting was 896 in training and test and the pretrained weights provided were used.

All trained models were tested on the Klinikum test-set. The results of the baseline models are displayed in table \ref{tab:baselinesTableResults}.

The results show, that CAD based synthetic data generally give better results than SCAN based data on this experiment. This is why we used SDRcad dataset as a baseline for all follow up experiments.

To investigate by how much the amount of real data can be reduced when used together with synthetic or MR data, while maintaining sufficient accuracy, experiments were conducted with a percentage distribution of the Klinikum training data. We decided to use the mosaic augmentation during these experiments as well and use all datasets as training data instead of a finetune experiment. We argue that the network can better learn relevant features while maintaining the advantages of the additional synthetic data when seeing a variation of all used datasets mixed together with mosaic augmentation as when only finetuning. During these experiments we decided to include the aforementioned green-channel augmentation on all trainings. Additionally the real data runs are trained with equal step size.

The follow up results with SDRcad, MR-DR as well as real data is shown in table \ref{tab:followUpResults}. While generally reaching worse results than the SDR based datasets, the MR dataset provides an additional boost in accuracy when used together with the SDRcad dataset.

These results show that when using synthetic data, MR data and all real data or only 15\% real data, the mAP accuracy could be improved further compared to when only using real data. Additionally it shows that when only using 15\% of real data the gap between the synthetic, MR and real data could be closed.

\begin{table}[ht]
	\small
	\caption{Results on Klinikum test-set for baseline trainings. *additional green-channel augmentation used}
	\centering
	\begin{tabular}{llllll}
		\multicolumn{5}{c}{\textbf{all}}\\
		\toprule	    
		Experiment     		& $mAP$  			& $mAP50$  			&  P 				&  R                   \\
		\toprule
		Klinikum train 	   	& 81.60				& 98.28 			& 88.99				& 98.73 \\
		\midrule
		DRscans      		& 46.06				& 75.51 			& 76.89				& 77.83 \\
		DRcad    	   		& 51.50				& 77.85 			& 76.40				& 81.56 \\
		SDRscans			& 65.52				& 87.12 			& 77.92				& 89.31 \\
		SDRcad  			& \textbf{67.44}	& \textbf{90.72} 	& \textbf{81.98}	& \textbf{92.02} \\
		MR-DR*  			& 60.94				& 82.29 			& 79.54				& 84.04 \\
		\bottomrule	    
	\end{tabular}
	\label{tab:baselinesTableResults}
\end{table}

\begin{table}
	\caption{Results on Klinikum test-set for follow up experiments. For these experiment we used the additional green-channel augmentation on all reported trainings.}
	\centering
	\begin{tabular}{llllll}
		\multicolumn{5}{c}{\textbf{all}}\\
		\toprule	    
		Experiment     		& $mAP$  			& $mAP50$  			&  P 				&  R                 \\
		\toprule
		\midrule
		Klinikum(100) 	   	& 81.95				& \textbf{98.57} 	& \textbf{88.88}	& \textbf{98.85} \\
		\midrule
		SDR+MR+\\real(100)  & \textbf{83.35}	& 98.14 			& 88.41				& 98.59 \\
		\midrule
		Klinikum(15)   		& 77.52				& 96.67 			& 87.83				& 97.37 \\
		\midrule
		SDR+MR+\\real(15)	& 80.05				& 96.92 			& 87.06				& 97.64 \\
		\midrule
		SDR+MR  			& 72.00				& 92.27 			& 83.81				& 94.07 \\
		\bottomrule	    
	\end{tabular}
	\label{tab:followUpResults}
\end{table}

\begin{figure}
	\centering
	\begin{tabular}{l c}
		\includegraphics[width=.22\textwidth,trim={15cm 5cm 25cm 0cm},clip]{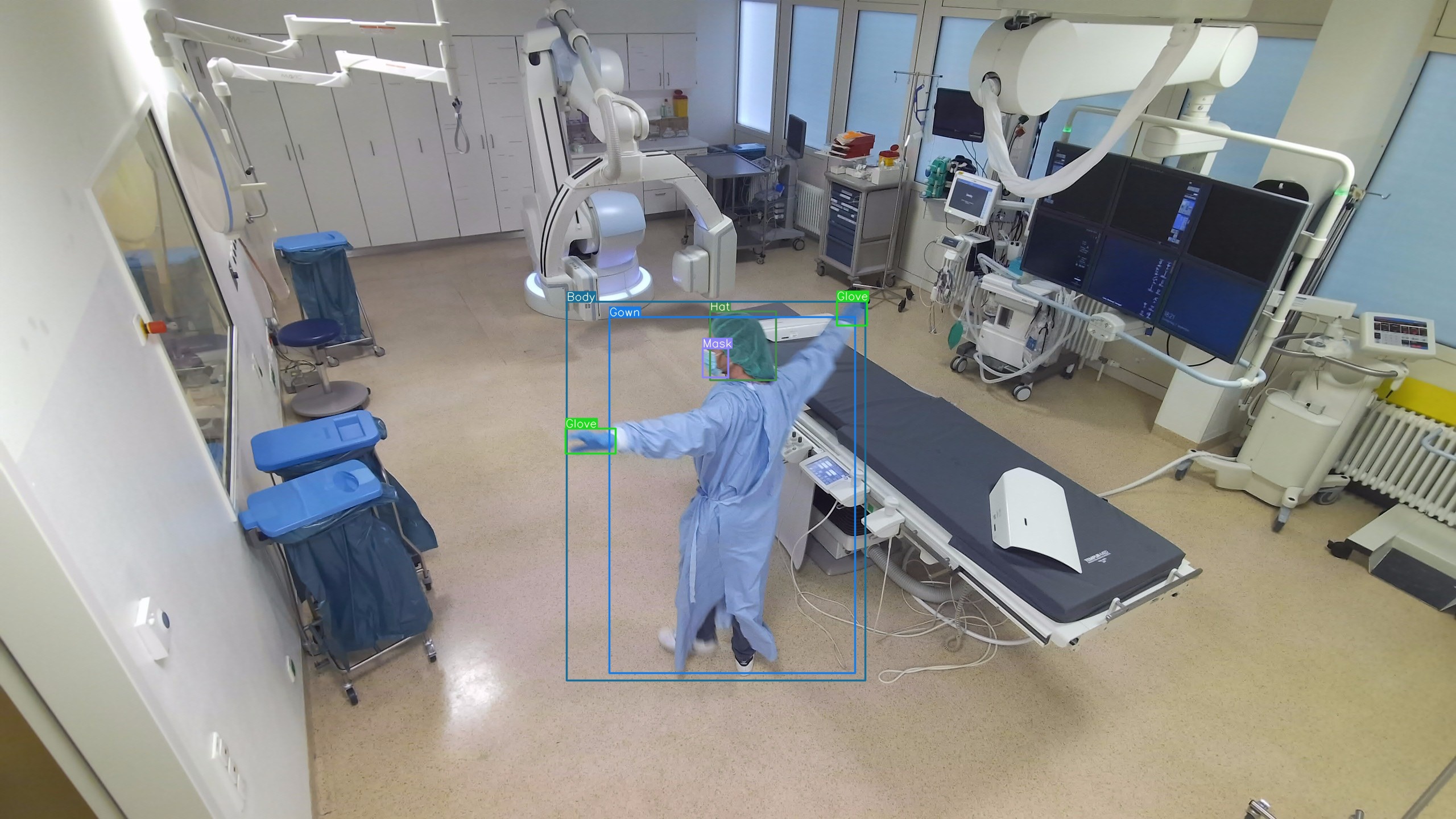}
		&
		\includegraphics[width=.22\textwidth,trim={20cm 8cm 25cm 2cm},clip]{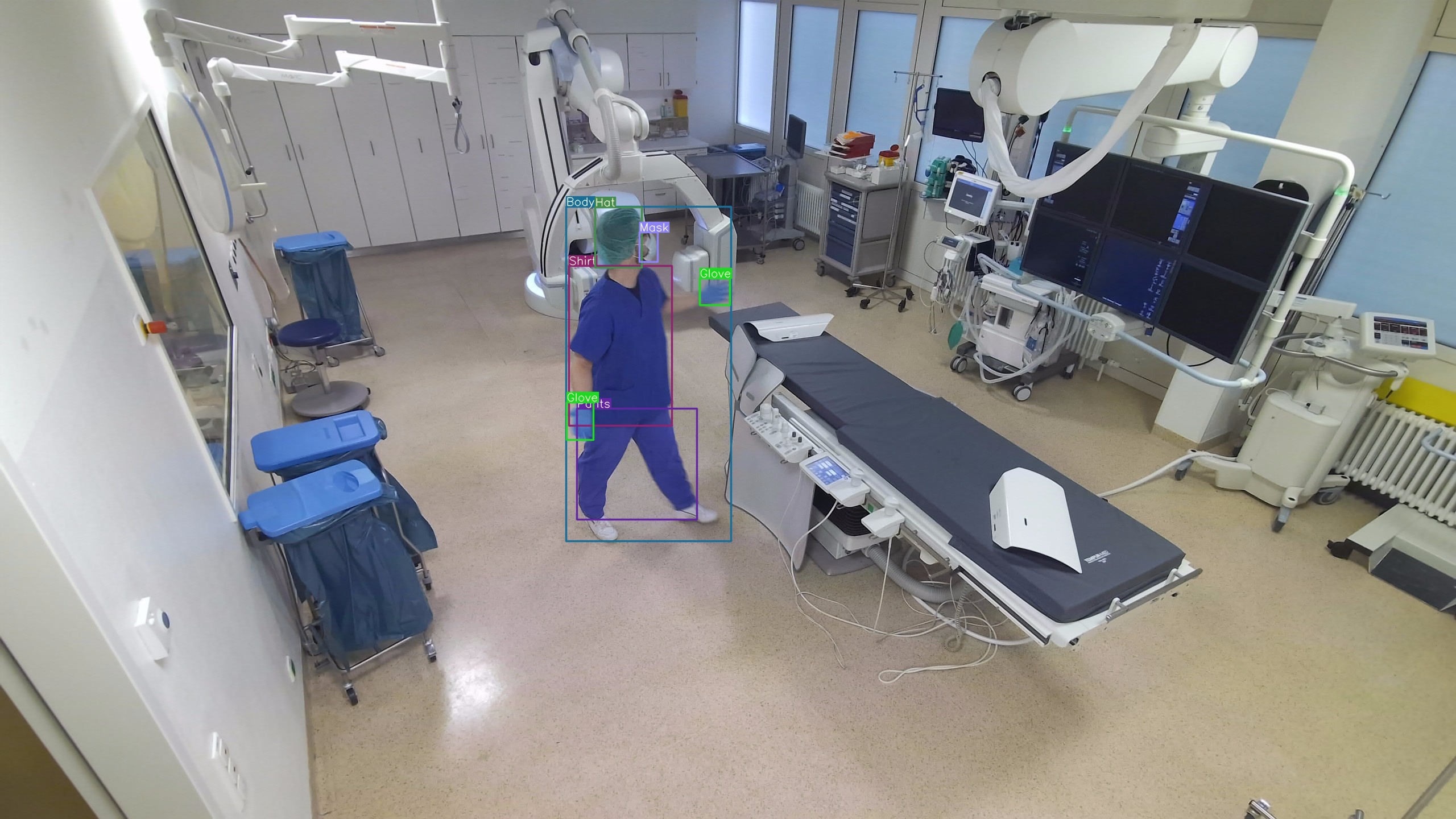} \\
		
		\includegraphics[width=.22\textwidth,trim={25cm 5cm 25cm 10cm},clip]{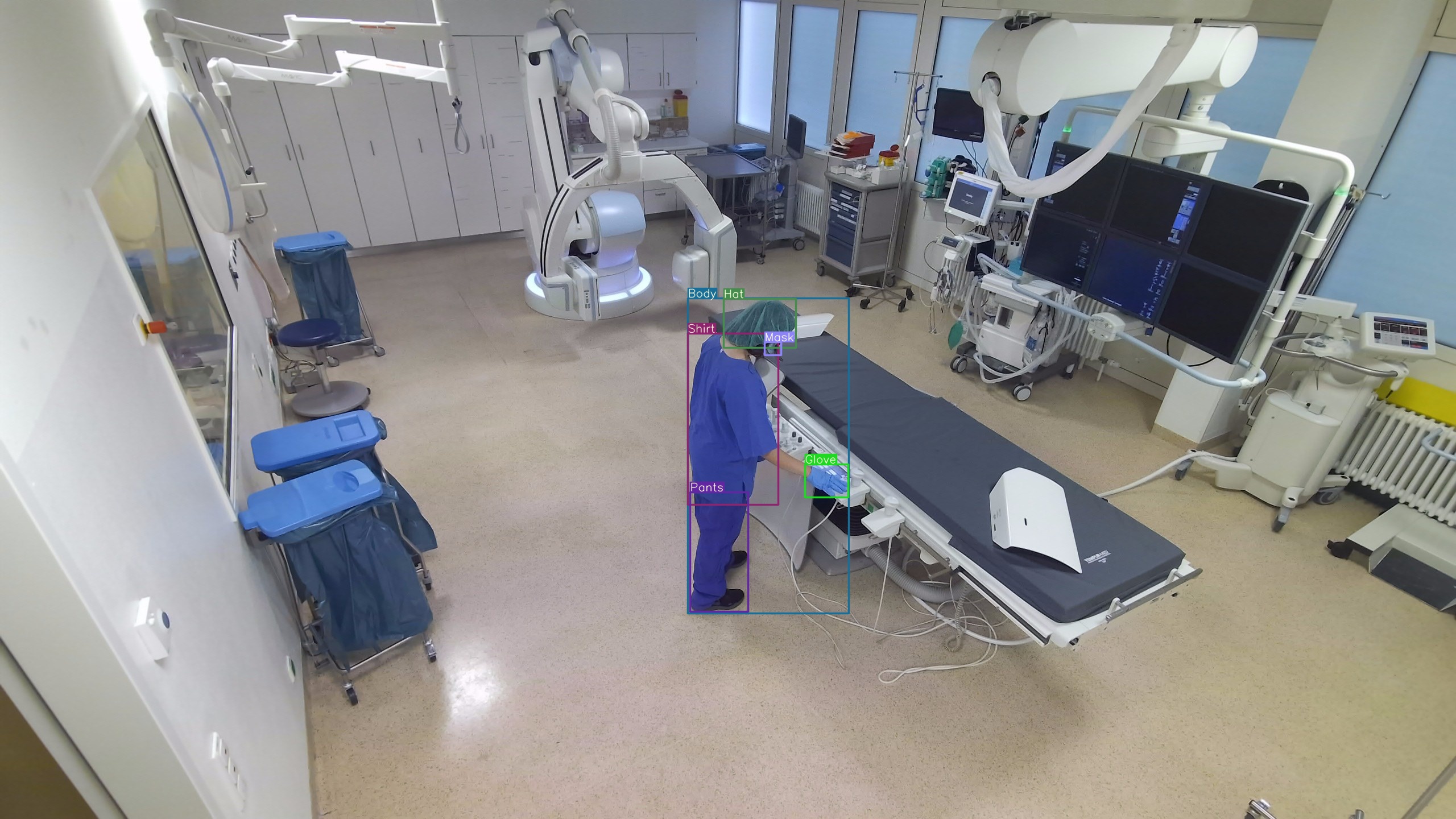}
		&
		\includegraphics[width=.22\textwidth,trim={20cm 5cm 25cm 5cm},clip]{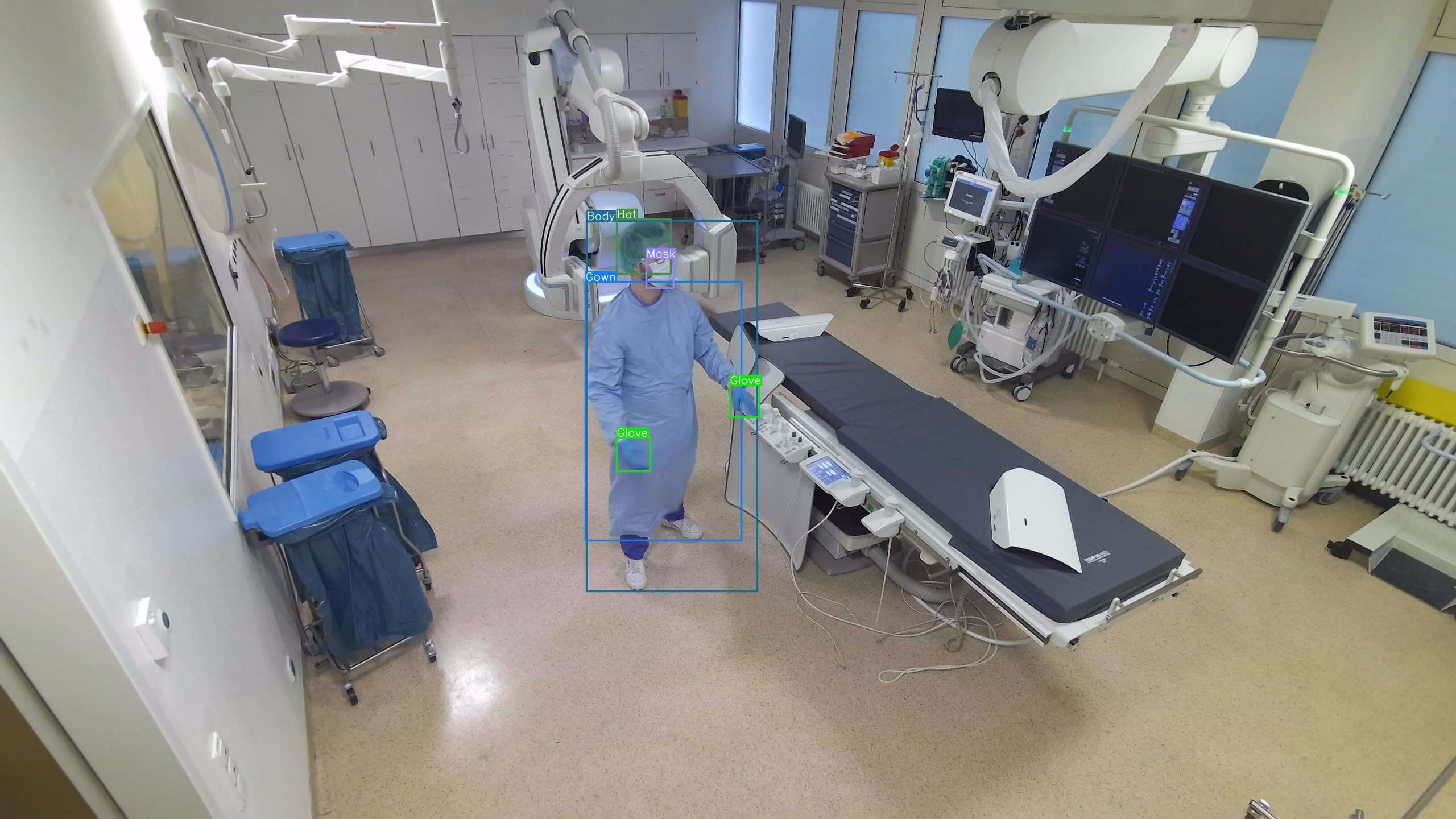} \\
	\end{tabular}	
	\caption{Inference results with SDR+MR+real(100) trained net. Confidence threshold of 0.4 and IoU-threshold of 0.5}
	\label{fig:bestScoreKlinikExp}
\end{figure}

Inference result images of SDR+MR+real(100) can be seen in figure \ref{fig:bestScoreKlinikExp}. Here we used a slightly higher confidence threshold of 0.4 and and IoU-threshold of 0.5.


\section{Conclusion}
\label{sec:conclusion}

We were able to show that the use of SMPL models together with scanned or designed medical clothing is a suitable method for modeling heathcare professionals for AI questions in the intervention space using the example of medical clothing detection.
During our experiments we found out that the designed clothing generally performed better on our test dataset than the 3D scanned cloths. This result surprised us, as we expected the potentially more accurate textures of the 3D scan to have a positive impact on detection rates. However, according to the results, it cannot be ruled out that artifacts in the rendering pipeline or pre-processing pipeline that we did not detect have an influence on this. In order to make a final statement about the potential of 3D scanned clothing for the modeling of health professionals, further experiments should be conducted.
Using Mixed-Reality data together with the synthetic data closed the gap further and while the margin is quite small, we could show that when using synthetic, mixed reality and 15\% real data the remaining gap towards 100\% real data could be nearly closed.

Finally, it can be stated that the presented modeling of health professionals is a promising methodology to address the challenge of missing datasets from medical intervention rooms. We will further investigate it on various task around the medical field.

\section{Acknowledgements}
This research project is part of the Research Campus M2OLIE and funded by the German Federal Ministry of Education and Research (BMBF) within the Framework “Forschungscampus – public-private partnership for Innovations” und the funding code 13GW0389C.

Disclaimer: The methods and information presented in this work are based on research and are not commercially available.

\bibliographystyle{unsrt}  
\bibliography{medClothPaper}  

\section{Appendix}
\label{sec:Appendix}

In this appendix chapter, the full table results of the baseline experiment and the follow experiments are provided. Addtionally to the all category we report the results for all detection classes (Body, Gown, Shirt, Pants, Hat, Mask, Glove). 

Table \ref{tab:fullbaselinesTableResults} shows the results of the baseline experiments and table \ref{tab:fullFollowUpTableResults} shows the results of the follow up experiments.

\begin{table*}[ht]
	\small
	\caption{results on Klinikum test-set for baseline trainings. *additional green-channel augmentation used}
	\centering
	\begin{tabular}{llllll}
		\toprule	    
		Experiment     		& $mAP$  & $mAP50$  &  P &  R                   \\
		\toprule
		\multicolumn{5}{c}{\textbf{all}}\\
		Klinikum train 	   	& 81.60							   & 98.28 						& 88.99				& 98.73 \\
		\midrule
		DRscans      		& 46.06							   & 75.51 						& 76.89				& 77.83 \\
		DRcad    	   		& 51.50							   & 77.85 						& 76.40				& 81.56 \\
		SDRscans			& 65.52							   & 87.12 						& 77.92				& 89.31 \\
		SDRcad  			& \textbf{67.44}				   & \textbf{90.72} 			& \textbf{81.98}	& \textbf{92.02} \\
		MR-DR*  			& 60.94							   & 82.29 						& 79.54				& 84.04 \\
		\bottomrule	    
		\multicolumn{5}{c}{\textbf{Body}}                   \\
		Klinikum train 	   	& 97.02							   & 99.49 						& 88.03				& 100 \\
		\midrule
		DRscans      		& 75.24							   & 96.26 						& 84.57				& 96.07 \\
		DRcad    	   		& 77.05							   & 98.39 						& 76.76				& 98.79 \\
		SDRscans			& 92.47							   & \textbf{99.50} 			& \textbf{89.95}	& \textbf{100} \\
		SDRcad  			& \textbf{92.88}							   & \textbf{99.50} 			& 85.09				& \textbf{100} \\
		MR-DR*  				& 80.88							   & 95.29 						& 82.55				& 95.77 \\
		\bottomrule	    
		\multicolumn{5}{c}{\textbf{Gown}}                   \\
		Klinikum train 	   	& 93.41							   & 99.5 						& 100				& 100 \\
		\midrule
		DRscans      		& 66.16							   & 87.5 						& 100				& 87.22 \\
		DRcad    	   		& 78.96							   & 94.17 						& 87.41				& 93.99 \\
		SDRscans			& 79.22							   & 94.77 						& 87.25				& 97.74 \\
		SDRcad  			& \textbf{85.86}				   & \textbf{99.49} 			& 99.25				& \textbf{99.25} \\
		MR-DR*  				& 50.18							   & 57.5 						& \textbf{100}		& 57.14 \\
		\bottomrule	    
		\multicolumn{5}{c}{\textbf{Shirt}}                   \\
		Klinikum train 	   	& 94.28							   & 99.47 						& 55.03				& \textbf{99.49} \\
		\midrule
		DRscans      		& 81.17							   & 98.19 						& \textbf{59.45}	& 98.48 \\
		DRcad    	   		& 82.82							   & 97.97 						& 56.85				& 98.48 \\
		SDRscans			& 84.72							   & 95.28 						& 56.76				& 95.45 \\
		SDRcad  			& 92.22							   & \textbf{99.41}				& 55.65				& \textbf{99.49} \\
		MR-DR*  				& 91.41							   & 97.72 						& 44.85				& 98.99 \\
		\bottomrule	    
		\multicolumn{5}{c}{\textbf{Pants}}                   \\
		Klinikum train 	   	& 86.81							   & 98.5 						& 98.98				& 98.48\\
		\midrule
		DRscans      		& 34.13							   & 88.44 						& 75.21				& 90.40 \\
		DRcad    	   		& 48.14							   & 92.22 						& \textbf{92.89}	& 92.42 \\
		SDRscans			& 54.45							   & 91.31 						& 90.05				& 91.41 \\
		SDRcad  			& 55.32							   & \textbf{95.61}				& 91.83				& \textbf{96.46} \\
		MR-DR*  				& 77.17							   & 92.26 						& 78.88				& 92.42 \\
		\bottomrule    
		\multicolumn{5}{c}{\textbf{Hat}}                   \\
		Klinikum train 	   	& 85.62							   & 99.27 						& 98				& 100 \\
		\midrule
		DRscans      		& 44.71							   & 78.73 						& 88.27				& 80.61 \\
		DRcad    	   		& 54.99							   & 94.41 						& \textbf{97.89}	& 94.90 \\
		SDRscans			& 78.44							   & 95.59 						& 67.94				& 99.49 \\
		SDRcad  			& \textbf{79.42}				   & 97.46 						& 97.45				& 97.45 \\
		MR-DR* 				& 70.51							   & 98.24 						& 95.57				& 98.98 \\
		\bottomrule	    
		\multicolumn{5}{c}{\textbf{Mask}}                   \\
		Klinikum train 	   	& 53.66							   & 95.85 						& 88.8				& 96.52 \\
		\midrule
		DRscans      		& 8.58							   & 44.07 						& 61.42				& 52.61 \\
		DRcad    	   		& 6.09							   & 29.59 						& 50.95				& 46.52 \\
		SDRscans			& \textbf{28.33}				   & 63.54 						& 78.77				& 66.09 \\
		SDRcad  			& 25.24							   & 71.91 						& 69.29				& 76.52 \\
		MR-DR*  				& 25.33							   & 73.75 						& \textbf{82.65}	& 78.70 \\
		\bottomrule	    
		\multicolumn{5}{c}{\textbf{Glove}}                   \\
		Klinikum train 	   	& 60.43							   & 95.88 						& 94.10				& 96.59 \\
		\midrule
		DRscans      		& 12.40							   & 35.36 						& 69.33				& 39.39 \\
		DRcad    	   		& 12.44							   & 38.17 						& 72.02				& 45.83 \\
		SDRscans			& 40.99							   & 69.88 						& 74.72				& \textbf{75} \\
		SDRcad  			& \textbf{41.10}				   & \textbf{71.63} 			& \textbf{75.29}	& \textbf{75} \\
		MR-DR*  			& 31.09							   & 61.26 						& 72.31				& 66.29 \\
		\bottomrule
	\end{tabular}
	\label{tab:fullbaselinesTableResults}
\end{table*}

\begin{table*}[ht]
	\small
	\caption{results on Klinikum test-set for follow up experiments. Green-channel augmentation used on all experiments.}
	\centering
	\begin{tabular}{llllll}
		\toprule	    
		Experiment     		& $mAP$  			& $mAP50$  			&  P 				&  R                 \\
		\toprule
		\multicolumn{5}{c}{\textbf{all}}\\
		\midrule
		Klinikum(100) 	   	& 81.95				& \textbf{98.57} 	& \textbf{88.88}	& \textbf{98.85} \\
		SDR+MR+real(100)  	& \textbf{83.35}	& 98.14 			& 88.41				& 98.59 \\
		Klinikum(15)   		& 77.52				& 96.67 			& 87.83				& 97.37 \\
		SDR+MR+real(15)		& 80.05				& 96.92 			& 87.06				& 97.64 \\
		SDR+MR  			& 72.00				& 92.27 			& 83.81				& 94.07 \\
		\bottomrule	    
		\multicolumn{5}{c}{\textbf{Body}}                   \\
		\midrule
		Klinikum(100) 	   	& 97.13				& \textbf{99.5}		& 87.33				& \textbf{100} \\
		SDR+MR+real(100)  	& \textbf{97.93}	& \textbf{99.5} 	& 86.42				& \textbf{100} \\
		Klinikum(15)   		& 94.24				& \textbf{99.5} 	& \textbf{92.46}	& \textbf{100} \\
		SDR+MR+real(15)		& 96.84				& \textbf{99.5} 	& 86.2				& \textbf{100} \\
		SDR+MR   			& 93.83				& 99.47 			& 87.00				& 99.1 \\
		\bottomrule	    
		\multicolumn{5}{c}{\textbf{Gown}}                   \\
		\midrule
		Klinikum(100) 	   	& 93.29				& \textbf{99.5}	 	& 99.25				& \textbf{100} \\
		SDR+MR+real(100)  	& \textbf{96.30}	& \textbf{99.5}		& \textbf{100}		& \textbf{100} \\
		Klinikum(15)   		& 87.88				& \textbf{99.5}		& \textbf{100}		& \textbf{100} \\
		SDR+MR+real(15)		& 95.57				& \textbf{99.5}		& 99.25				& \textbf{100} \\
		SDR+MR   			& 85.11				& 95.47 			& 99.22				& 95.49 \\
		\bottomrule	    
		\multicolumn{5}{c}{\textbf{Shirt}}                   \\
		\midrule
		Klinikum(100) 	   	& 94.51				& \textbf{99.46} 	& 55.49				& \textbf{99.5} \\
		SDR+MR+real(100)  	& \textbf{96.4}		& 99.37 			& 57.10				& \textbf{99.5} \\
		Klinikum(15)   		& 90.43				& 98.07 			& 57.18				& 98.5 \\
		SDR+MR+real(15)		& 95.62				& 99.38 			& \textbf{60.43}	& \textbf{99.5} \\
		SDR+MR   			& 94.17				& 99.31 			& 54.42				& \textbf{99.5} \\
		\bottomrule	    
		\multicolumn{5}{c}{\textbf{Pants}}                   \\
		\midrule
		Klinikum(100) 	   	& 87.16				& \textbf{99.39} 	& \textbf{99.49}	& \textbf{98.99} \\
		SDR+MR+real(100)  	& \textbf{88.22}	& 98.49 			& \textbf{99.49}	& 98.48 \\
		Klinikum(15)   		& 81.70				& 97.40 			& 95.12				& 98.48 \\
		SDR+MR+real(15)		& 85.78				& 97.5	 			& 97.97				& 97.47 \\
		SDR+MR  			& 80.22				& 95.26				& 91.30				& 95.45 \\
		\bottomrule	    
		\multicolumn{5}{c}{\textbf{Hat}}                   \\
		\midrule
		Klinikum(100) 	   	& 84.77				& 99.29 			& 97.51				& \textbf{100} \\
		SDR+MR+real(100)  	& \textbf{86.90}	& \textbf{99.3} 	& \textbf{98}		& \textbf{100} \\
		Klinikum(15)   		& 83.77				& 98.59 			& 96.06				& 99.5 \\
		SDR+MR+real(15)		& 84.91				& 99.27 			& 97.51				& \textbf{100} \\
		SDR+MR  			& 77.6				& 99.22 			& 97.51				& \textbf{100} \\
		\bottomrule	    
		\multicolumn{5}{c}{\textbf{Mask}}                   \\
		\midrule
		Klinikum(100) 	   	& \textbf{55.42}	& \textbf{95.70}	& \textbf{91.32}	& \textbf{96.09} \\
		SDR+MR+real(100)  	& 54.70				& 93.84 			& 88.62				& 94.78 \\
		Klinikum(15)   		& 50.30				& 93.17 			& 85.66				& 93.48 \\
		SDR+MR+real(15)		& 44.28				& 90.95 			& 82.81				& 92.17 \\
		SDR+MR  			& 27.61				& 78.89 			& 76.36				& 85.65 \\
		\bottomrule	    
		\multicolumn{5}{c}{\textbf{Glove}}                   \\
		\midrule
		Klinikum(100) 	   	& 61.39				& \textbf{97.11}	& \textbf{91.79}	& \textbf{97.35} \\
		SDR+MR+real(100)  	& \textbf{62.98}	& 97.0 				& 89.24				& \textbf{97.35} \\
		Klinikum(15)   		& 54.29				& 90.46 			& 88.32				& 91.67 \\
		SDR+MR+real(15)		& 57.32				& 92.32 			& 85.27				& 94.32 \\
		SDR+MR  			& 45.46				& 78.29 			& 80.88				& 83.33 \\
		\bottomrule	    
	\end{tabular}
	\label{tab:fullFollowUpTableResults}
\end{table*}

\end{document}